\documentclass[runningheads]{llncs}

\usepackage{eccv}

\usepackage{multirow}
\usepackage{nicematrix}
\usepackage{amssymb}
\usepackage{xcolor}
\usepackage{wrapfig2}

\definecolor{cvprblue}{rgb}{0.21,0.49,0.74}
\definecolor{lightgray}{gray}{0.9}
\definecolor{lightpurple}{rgb}{0.9,0.9,1.0}
\usepackage{eccvabbrv}

\usepackage{graphicx}
\usepackage{booktabs}

\usepackage[accsupp]{axessibility}  

\usepackage{hyperref}

\usepackage{orcidlink}

\begin{document}

\title{IRG-MotionLLM: Interleaving Motion Generation, Assessment and Refinement for Text-to-Motion Generation} 

\titlerunning{IRG-MotionLLM}

\author{Yuan-Ming Li\inst{1,2,4}\and
Qize Yang\inst{2} \and
Nan Lei\inst{1,4}\and
Shenghao Fu\inst{1,4} \and
Ling-An Zeng\inst{1,4} \and
Jian-Fang Hu\inst{1,4} \and
Xihan Wei\inst{2} \and
Wei-Shi Zheng\inst{1,3,4,\dagger}}

\authorrunning{Y.M. Li et al.}

\institute{
$^{1}$ Sun Yat-sen University;
$^{2}$ Tongyi Lab, Alibaba Group;\\
$^{3}$ Shenzhen Loop Area Institute;
$^{4}$ Key Laboratory of Machine Intelligence and Advanced Computing, Ministry of Education, China
}

\maketitle

{\let\thefootnote\relax\footnotetext{
\scriptsize 
{$\dagger$}: Corresponding author. \\
\textbf{Emails:} liym266@mail2.sysu.edu.cn; wszheng@ieee.org. 
}}

\vspace{-0.2cm}
\begin{abstract}
  Recent advances in motion-aware large language models have shown remarkable promise for jointly learning motion understanding and generation knowledge. However, these models typically treat understanding and generation separately, limiting the mutual benefits that could arise from interactive feedback between tasks. In this work, we reveal that motion assessment and refinement tasks can act as crucial bridges to enable knowledge flow from motion understanding to generation. Specifically, we propose Interleaved Reasoning for Motion Generation (IRMoGen), a novel paradigm that tightly couples motion generation with assessment and refinement through iterative text-motion dialogue. To realize this, we introduce IRG-MotionLLM, the first model that seamlessly interleaves motion generation, assessment, and refinement to improve the alignment between generated motion and goal text. IRG-MotionLLM is developed progressively with a novel three-stage training scheme, initializing and subsequently enhancing native IRMoGen capabilities. To facilitate this development, we construct an automated data engine to synthesize interleaved reasoning annotations from existing text-motion datasets. Extensive experiments demonstrate the properties brought by IRMoGen training, and the advanced cross-benchmark and cross-evaluator performance of IRG-MotionLLM. 
  Code and models are available at \textcolor{purple}{\url{https://github.com/HumanMLLM/IRG-MotionLLM}}.
  
\end{abstract}

\vspace{-0.2cm}
\begin{figure}[t]
  \centering
   \includegraphics[width=1\linewidth]{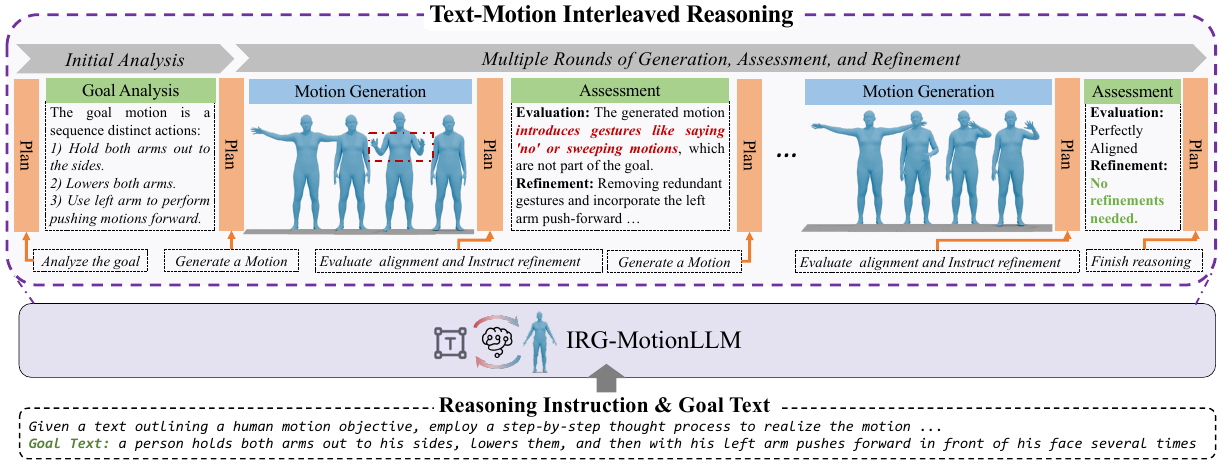}
	\vspace{-0.7cm}
   \caption{\textbf{Illustration of our proposed IRG-MotionLLM.}Given a goal text and reasoning instruction as input, our IRG-MotionLLM is able to \textbf{perform Text-Motion Interleaved Reasoning} until a satisfactory motion is generated. The reasoning process includes multiple moves, \ie, an initial analysis on the goal, followed by multiple rounds of motion \emph{Generation}, \emph{Assessment}, and \emph{Refinement}. The model can also adaptively make plans for the next move.}
   \label{fig:pipeline}
   \vspace{-0.50cm}
\end{figure}

\vspace{-0.4cm}
\section{Introduction}
\label{sec:intro}
\vspace{-0.2cm}

Text-to-motion generation has emerged as a cornerstone problem in computer vision, with transformative applications in multiple human–robot interaction \cite{ji2025towards,huang2026learning,yin2026spatiotemporal,lei2026physigen}, and building immersive virtual 3D world\cite{cao2025uni3c,hou2026egomotion,wu2026plenodium}. 
To promote the development of this area, numerous methods \cite{mogen_survey1,mogen_survey2} have been proposed to generate motions with a higher naturalness and alignment level with the goal.

Inspired by recent advances in multimodal large language models \cite{wang2023image,huang2023language,blip,blip2,lin2025pancap,song2025towards}, efforts in Motion-aware LLM (MoLLM) \cite{motiongpt, motiongpt2,motionchain,mg-motionllm} propose to jointly train motion understanding and generation tasks with a single foundation model. While joint training provides a versatile basis for various motion-related tasks (\eg, text-to-motion generation, motion-to-text caption), a critical limitation remains: motion understanding and generation are predominantly modeled in isolation without any explicit interactions, which probably leads to the marginal improvement from complementary learning on various tasks.

Most recently, emerging works \cite{Uni-cot, irg, chern2025thinking} propose to enhance the generation performance of the multimodal model by tightly coupling generation and understanding via \emph{interleaved reasoning} across text and images.
This observation naturally motivates us to ponder a question: \textbf{\emph{Can we further improve MoLLM's text-to-motion generation capability by explicitly linking motion-related tasks through text-motion interleaved reasoning?}}
 
In this work, we provide the first affirmative answer to this question by introducing Interleaved Reasoning for Motion Generation (IRMoGen), a novel paradigm that bridges motion understanding and generation via iterative text-motion dialogue. We argue that achieving this vision is non-trivial and entails three progressive challenges: 

(1) \textbf{\emph{Task Gap}}: Despite sharing latent representations, basic motion generation (\eg, text-to-motion) and understanding (\eg, motion caption) tasks remain disconnected in training and inference. There is no intermediate task that enables the model to leverage motion understanding to improve generation.
(2) \textbf{\emph{Training Scheme}}: Beyond the task gap, an advanced training strategy is required to initialize and further enhance the IRMoGen capability.
(3) \textbf{\emph{Data Absence}}: There is a lack of an effective data engine to generate text-motion interleaved reasoning annotations, which are necessary for training the model.

For the first challenge, we propose to employ text-motion \emph{assessment} and \emph{refinement} tasks to bridge the gap between motion understanding and generation. The former requires evaluation of the alignment between the given motion and the goal text and refinement instructions. The latter asks the model to refine the initial motion based on the goal text and refinement instructions. These two tasks enable us to establish an IRMoGen process by \emph{interleaving Motion Generation, Assessment, and Refinement}.

Based on this framework, we introduce IRG-MotionLLM, the first model designed to support native IRMoGen. 
As shown in \cref{fig:pipeline}, given a goal text and reasoning instruction as input, IRG-MotionLLM performs step-by-step reasoning, which contains initial goal analysis, generation, and multiple rounds of assessment and refinement until a satisfactory motion is generated.
Beginning with a pre-trained Motion-aware LLM \cite{motionagent} as the base model, we propose a three-stage training scheme to imbue and enhance the ability required for IRMoGen.

In the first \textbf{IRMoGen Initialization} stage, we fine-tune the model on eight atomic sub-tasks, including 4 \emph{basic} tasks (related to basic understanding and generation) and 4 \emph{improving} tasks (assessment- and refinement-related tasks to implicitly bridge motion understanding and generation). 
In the second \textbf{IRMoGen-CoT Learning} stage, we design a reasoning CoT template to explicitly connect various motion-related knowledge, and train the model to automatically make a plan to interweave initial goal analysis, motion generation, together with multi-round motion assessment and refinement. 
In the final \textbf{IRMoGen Reinforcing} stage, we employ GRPO \cite{deepseek-math} to further unlock the reasoning potential of the model. With the carefully designed reward functions, the model is able to freely explore multi-round IRMoGen and improve the final generated motion.

To facilitate our exploration, we further design an automated data engine to obtain IRMoGen annotations for the existing text-motion datasets (\eg, HumanML3D\cite{guo}, KIT-ML\cite{kit-ml}). With the help of a pre-trained motion encoder and LLM, we are able to assign prompt analysis, multiple negative motions with various alignment levels, together with the text-motion evaluation and refinement instructions for each text-motion pair in the dataset. 

Combining all of these, we conduct extensive experiments to showcase the properties and verify the effectiveness of our method. Our key findings include: 

\noindent - \textbf{\emph{Introducing motion assessment and refinement tasks largely benefits text-motion alignment.}} After Stage-1 training, IRG-MotionLLM not only achieves stable improvement on the text-to-motion generation task, but also outperforms existing methods on the motion-to-text caption task. 

\noindent - \textbf{\emph{Interweaving Generation, Assessment, and Refinement significantly improves alignment between generated motions and goal texts.}} This finding is consistent across all training stages. Surprisingly, our Stage-1 model emerges IRMoGen abilities even if it is not explicitly trained to perform IRMoGen. The Stage-2 and Stage-3 training bring further improvement. In addition, we observe the same trends as in \cite{guo2025deepseek} that the model responds with a longer reasoning process after RL tuning, and the alignment between generated motion and goal text benefits from more rounds of assessment and refinement.

\noindent - \textbf{\emph{IRG-MotionLLM clearly outperforms the base model and achieves advanced performance}}. The results are consistent across different benchmarks \cite{guo,kit-ml} and  different evaluators \cite{guo, mardm}.  

\noindent - \textbf{\emph{IRG-MotionLLM enables cross-model and cross-task synergies}}. Beyond performing the T2M task independently, our model can serve as a transferable proxy to enhance existing motion generator via an RLAIF manner \cite{atom}. Such emergent property has never been explored in previous works on MoLLM. We also show that our method benefits the adaptation to motion editing task \cite{motionfix}.

We will make the code and models available. We expect our exploration and findings will benefit the field of motion generation and MoLLM.

\section{Related Works}
\vspace{-0.2cm}

\noindent{\textbf{Text-Driven Human Motion Generation}} aims at generating 3D human motions based on text description. Most modeling paradigms in this field can be separated into three families, \ie, diffusion modeling \cite{mdm, motiondiffuse, mld, mardm, li2026motionhiflow,li2026moftss,zeng2025progressive}, mask token modeling \cite{mmm,momask,bamm,lamp,guo2025snapmogen,mhm}, and autoregressive modeling \cite{t2m-gpt, go-to-zero, scamo, being-m0}. Based on their success, some other works explore better motion representations \cite{mardm, jiang2025causal,meng2025absolute}, detailed text prompts \cite{finemogen,he2023semanticboost,yazdian2023motionscript}, RAG techniques \cite{remodiffuse,remogpt,rmd,remomask}, lightweight motion models \cite{light-t2m, motionmamba}, etc.
Notably, with the belief that motion generation and understanding knowledge are complementary, works on Motion-aware LLM \cite{motiongpt, motiongpt2,mg-motionllm,momug,motiongpt3,hou2025motionverse,egolm,m3gpt,being_m05,li2026llamo} propose to jointly learn multiple tasks with a single model. 
Despite providing versatile basis for various motion-centric tasks, existing works in this line suffer from a fundamental limitation: \emph{generation and understanding tasks are trained and executed in isolation without explicit cross-task interaction}, weakening the complementary role of understanding and generating knowledge.
MotionChain \cite{motionchain} and MotionAgent \cite{motionagent} perform motion understanding and generation sequentially, but focus on multi-turn conversation rather than enhancing generation performance via motion understanding. Besides, they either require heavy human intervention or an extra LLM to connect motion generation and caption.
Most recently, Motion-R1 \cite{motion-r1} explores the interaction between understanding and generation. However, it only performs motion-free text understanding before generation.
In contrast, our work is the first to explore the native interaction loop across motion generation, assessment, and refinement. 

\vspace{0.1cm}
\noindent{\textbf{Multi-modal Interleaved Reasoning}}:
Inspired by the success of text-based reasoning models \cite{openai-o1,guo2025deepseek,vision-r1,fu2025love}, multi-modal interleaved reasoning \cite{openai-o3,ir-github,gu2025thinkmorph} is proposed to solve complex problems, \ie, incorporating non-text modalities into multi-turn reasoning processes. In the vision generation domain, recent works \cite{t2i-r1,got,bagel} explore improving generation performance with text-based reasoning. Emerging research \cite{Uni-cot, irg, chern2025thinking, zhuo2025reflection,omnigen2} further shows that the generated image can better align with the text prompt via text-image interleaved reasoning, including self-assessment and refinement.
However, this property has never been observed in human motion generation. In this work, we take the first step to bring MoLLM to this frontier.

\begin{figure}[t]
  \centering
   \includegraphics[width=1\linewidth]{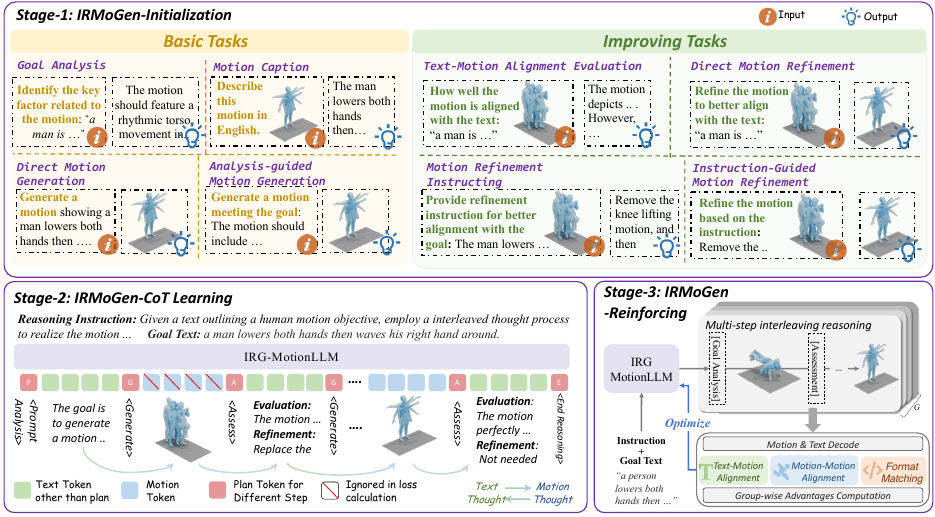}
	\vspace{-0.7cm}
   \caption{\textbf{A three-stage training scheme is proposed to build our IRG-MotionLLM}. \textbf{In the first stage (Upper)}, we endow the model with meta IRMoGen abilities via eight related tasks. \textbf{In the second stage (Lower-Left)}, we define an IRMoGen-CoT template and train a model to explicitly couple motion understanding and generation knowledge and perform native interleaved reasoning. \textbf{In the third stage (Lower-Right)}, we employ GRPO-based reinforcement learning to unleash the multi-round reasoning potential of the model.}
   \label{fig:training_scheme}
   \vspace{-0.3cm}
\end{figure}

\section{IRG-MotionLLM}
\vspace{-0.2cm}
In this work, we propose IRG-MotionLLM, the first motion-aware LLM that supports native interleaved reasoning across text and motion to enhance final generation performance. Our key idea is to interweave motion generation, assessment, and refinement into the IRMoGen reasoning process. 
To this end, we build our IRG-MotionLLM based on a pre-trained base model and propose a three-stage training scheme to initialize and further enhance the IRMoGen ability of the model. In the following sections, we first introduce the model pipeline and the base model in  \cref{sec:base_model}, then the Interleaving Reasoning Initialization (Stage-1) in \cref{sec:IR_init}, and finally the Interleaving Reasoning Enhancement (Stage-2 \& 3) in \cref{sec:IR_enhance}.

\vspace{-0.1cm}
\subsection{Pipeline \& Base Model}
\label{sec:base_model}
\vspace{-0.1cm}

IRG-MotionLLM shares the same architecture with previous works on MoLLM \cite{motiongpt,motionagent,mg-motionllm}, which consists of 
a Motion VQVAE, and an LLM.
Specifically, Motion VQVAE discretizes the motion sequence into \emph{K} distinct motion tokens. These motion tokens are mapped to a motion token vocabulary based on their indices.
We then extend the base LLM vocabulary with these motion-specific codes along with boundary tokens \textless Motion\textgreater and \textless /Motion\textgreater to delineate motion sequence spans within the text(\ie, K+2 new tokens). This setup allows the model to be trained with a standard next-token prediction objective.
During inference, the model can freely generate sequences that mix both text and motion tokens.
The generated motion can be directly extracted by identifying the boundary tokens.
With this general architecture, it will be convenient to build our IRG-MotionLLM upon a base model, which can be initialized with the pretrained weights or preliminary training strategies proposed in previous works. We provide more details in the Appendix.

\vspace{-0.1cm}
\subsection{IRMoGen Initialization (Stage-1)}
\label{sec:IR_init}
\vspace{-0.1cm}
As discussed in \cref{sec:intro}, although existing MoLLMs show versatility on various motion-related tasks (e.g., motion generation, caption, and prediction), these tasks can only be performed independently without interactions, limiting their ability to perform the IRMoGen as shown in \cref{fig:pipeline}. 
To solve this, we propose to initialize the interleaved reasoning ability with eight related tasks, which are divided into two categories, shown in the upper part of \cref{fig:training_scheme}:

\vspace{0.1cm}
{- \textbf{Basic Text-Motion Tasks}} require the model to perform basic translation between motion and text, including: 
(1) \emph{Motion Caption}: Describe what is happening in the motion; 
(2) \emph{Direct Motion Generation}: Generate motion based on the caption; 
(3) \emph{Prompt Analysis}: Analyze the key factors that need to be focused on to generate the target motion; 
(4) \emph{Analysis-guided Motion Generation}: Generate motion based on the results of prompt analysis.
The former two tasks are widely used for training MoLLM. Additionally, we add the latter two tasks to preserve the language reasoning ability and enhance text-motion alignment.

\vspace{0.1cm}
{- \textbf{Improving Tasks}} ask the model to perform deeper reasoning across text and motion, including:
(1) \emph{Text-Motion Alignment Evaluation}: Analyze how well the given motion is aligned with the goal text;
(2) \emph{Motion Refinement Instructing}: Provide a refinement instruction based on the given motion and the goal text to improve alignment.
(3) \emph{Direct Motion Refinement}: Refine the initial motion simply based on the goal text.
(4) \emph{Instruction-guided Motion Refinement}: Refine the given motion based on both the goal text and the refinement instructions. 
These tasks explicitly require the model to mine {the} differences between motions and texts, helping the model to construct a deeper alignment between these two modalities.

\vspace{0.1cm}
\noindent{\textbf{Training}}: 
By constructing task-specific instruction templates and annotations, we conduct supervised fine-tuning on the base model and obtain our Stage-1 model $\mathcal{F}_{s1}$. The annotation construction process and instruction templates are detailed in \cref{sec:data_engine} and the Appendix.

\vspace{0.1cm}
\noindent{\textbf{Inference}}: 
Similar to previous MoLLMs, $\mathcal{F}_{s1}$ can be instructed to perform all learned tasks.
More importantly, $\mathcal{F}_{s1}$ emerges IRMoGen ability as we are able to manually perform multi-step text-motion interleaved reasoning by sequentially using prompts for \emph{generation}, \emph{assessment}, and \emph{refinement} tasks together with the outputs from the previous step. See more details in the Appendix.

\vspace{-0.1cm}
\subsection{IRMoGen Enhancement}
\label{sec:IR_enhance}
\vspace{-0.1cm}
While the first training stage preliminarily endows motion with IRMoGen ability, it still requires human intervention.
Ideally, as shown in \cref{fig:pipeline}, given a text description for a goal motion, the model should be able to natively think step-by-step to interweave generation, assessment, and refinement until a satisfactory motion is obtained. During this process, the model needs to plan to decide which action to take.
We also believe that explicitly training the model to link the automated knowledge learned in Stage-1 can further enhance the interleaved reasoning ability and further benefit generation performance.

To this end, we first define an IRMoGen-CoT template to explicitly link knowledge learned in Stage-1. Subsequently, we propose IRMoGen-CoT learning and IRMoGen Reinforcing to enable and further unleash the structured IRMoGen ability of IRG-MotionLLM.

\begin{wrapfigure}{r}{0.5\textwidth}
  \centering
	\vspace{-0.7cm}
   \includegraphics[width=1\linewidth]{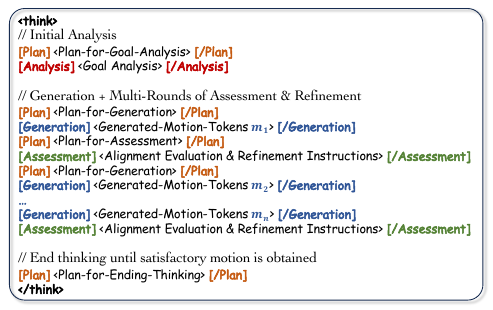}
	\vspace{-0.7cm}
   \caption{The IRMoGen-CoT Template.}
   \label{fig:cot_template}
   \vspace{-0.5cm}
\end{wrapfigure}

\vspace{0.1cm}
\noindent{\textbf{IRMoGen-CoT Template}}: To enable structured IRMoGen, we design a multi-step CoT template to accommodate Motion Generation, Assessment, and Refinement.
As shown in \cref{fig:cot_template}, the template begins with goal-directed planning and analysis, followed by an iterative loop that tightly interleaves motion token generation with alignment assessment and targeted refinement until a satisfactory motion is obtained (\ie, the assessment results indicate \emph{``no refinement is needed''}). In addition, each step is connected with a pre-defined \emph{plan making} template (\eg, ``I need to generate a motion based on previous analysis.'' for \textless Plan-for-Generation\textgreater).
Note that this CoT definition results in a multi-time generation reasoning trajectory. Suppose a trajectory contains $n$-time generations (\ie, $\{m_1, m_2, ..., m_n\}$), we note $\{m_1,..,m_{n-1}\}$ as \emph{intermediate motions} and $m_n$ as \emph{final motion}.
More details about the CoT template and data example are shown in the Appendix.

\vspace{0.1cm}
\noindent{\textbf{IRMoGen-CoT Learning (Stage-2)}}: 
Enabled by the data engine introduced in \cref{sec:data_engine}, we organize dynamic reasoning trajectories to train $\mathcal{F}_{s1}$ to perform native interleaved reasoning, and obtain our Stage-2 model  $\mathcal{F}_{s2}$ (shown in the lower-left part of \cref{fig:training_scheme}).
Each training data set is designed to match the IRMoGen-CoT. Additionally, the \emph{intermediate motions} $\{m_1, ..., m_{n-1}\}$ are misaligned with the goal text but show progressive improvement (\ie, $m_2$ is more aligned with the goal than $m_1$), while the \emph{final motion} $m_n$ is the ground-truth motion paired with the goal text.

While adopting vanilla next-token prediction objective can optimize the model to learn reasoning format, training the model to generate incorrect \emph{intermediate motions} will significantly break the already learned alignment between motion and text.
To address this, during training in this stage, we adopt an \textbf{\emph{Ignore Incorrect}} strategy to mask the losses and gradients calculated on those incorrect intermediate motion tokens.

After training the model on the IRMoGen-CoT dataset, $\mathcal{F}_{s2}$ is able to natively conduct interleaved reasoning across text and motion. However, the model tends to stop thinking at an early step (\ie, more than 70\% reasoning processes on test samples contain only one-round generation without any refinement), suggesting the multi-round assessment and refinement ability may be limited. This observation motivates us to seek further improvement in Stage-3 training.

\vspace{0.1cm}
\noindent{\textbf{IRMoGen Reinforcing (Stage-3)}}: 
To further unleash the multi-round interleaved reasoning potential, inspired by the success of reinforcement finetuning on (vision-)language tasks \cite{guo2025deepseek, vision-r1}, we conduct a GRPO\cite{deepseek-math}-based IRMoGen Reinforcing stage on $\mathcal{F}_{s2}$ and obtain our stage-3 model $\mathcal{F}_{s3}$.

Specifically, as shown in the lower right part of \cref{fig:training_scheme}, taking the reasoning instruction and motion text as input $q$, a group of $G$ IRMoGen trajectories $\{o_1,o_2,...,o_G\}$ are sampled from the model. After that, we assign each trajectory a scalar reward $r(o_i)$. 
Specifically, we employ reward functions comprising the following components:

{\textbf{1) Format Reward}} to encourage structured reasoning and easy answer extraction, responses must follow the IRMoGen-CoT template. Formally:

\begin{equation}
r_{\text{form}}(o) = 
\begin{cases}
1, & \text{if } o \text{ follows the required format}, \\
0, & \text{otherwise}.
\end{cases}
\end{equation}

{\textbf{2) Text-Motion Alignment Reward}} to improve alignment between the final generated motions in the reasoning trajectory and the goal text prompts: 

\begin{equation}
r_{tm}(t, \mathbf{m_{final,i}}) = - \|E_t(t) - E_m(\mathbf{m_{final,i}})\|^2,
\end{equation}

\noindent where $t$ indicates goal text prompt, $\mathbf{m_{final,i}}$ denotes the final generated motion in the reasoning trajectory $o_i$. $E_t$ and $E_m$ indicate the paired text and motion encoders. Following previous works \cite{motion-r1, motionrl, instructmotion-arxiv}, $E_t$ and $E_m$ are chosen from paired text and motion encoders in \cite{guo}.

{\textbf{3) Motion-Motion Alignment Reward}} to improve alignment between the final generated motions in the reasoning trajectory and the ground-truth motion. Formally: 

\begin{equation}
r_{mm}(\mathbf{m_{gt}}, \mathbf{m_{final,i}}) = - \|E_m(\mathbf{m_{gt}}) - E_m(\mathbf{m_{final,i}})\|^2,
\end{equation}
where $\mathbf{m_{gt}}$ indicates ground-truth motion sequence. Note that $r_{mm}$ will only be calculated for those data with a motion-text pair; otherwise set to 0.

Each of these rewards is pre-normed across the group into the same scale, and summed to calculate $r(o_i)$. After that, we employ the standard GRPO objective \cite{deepseek-math} to optimize our IRG-MotionLLM. See more details in the Appendix.

\section{Data Engine}
\label{sec:data_engine}
\vspace{-0.2cm}

To enable our exploration, we design a highly automated annotation pipeline to augment existing motion-text datasets $\textit{D}$ based on a pre-trained motion encoder \cite{tmr} and frontier LLMs. With this annotation pipeline, for each text-motion pair in the dataset, we are able to assign it with: (1) Detailed analysis on the generation goal; (2) Negative text-motion pairs with various alignment levels, together with the text-motion alignment evaluations and refinement instructions.

\vspace{0.1cm}
\noindent{\textbf{Detailed Goal Analysis}}: For each text prompt $t_i$ in dataset $\textit{D}$, we adopt GPT-4o \cite{gpt-4o} to \emph{identify} the key factors to generate the motion aligned with $t_i$. 

\begin{wrapfigure}{r}{0.5\textwidth}
  \centering
   \vspace{-0.5cm} 
   \includegraphics[width=1\linewidth]{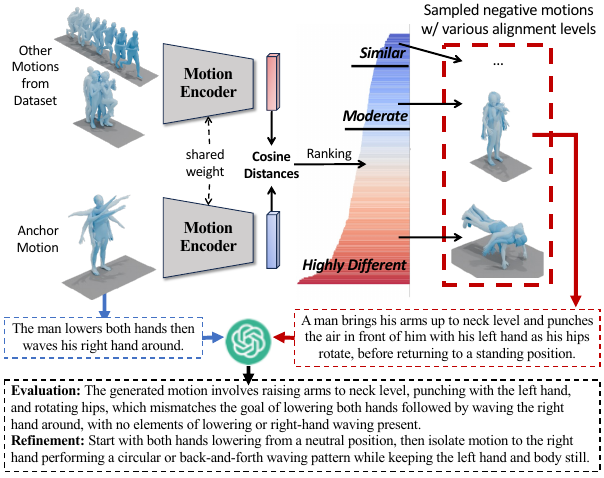}
   \vspace{-0.7cm}
   \caption{Negative text-motion pairs selection and annotation pipeline.}
   \label{fig:date_engine}
   \vspace{-0.3cm} 
\end{wrapfigure}

\vspace{0.1cm}
\noindent{\textbf{Negative Text-Motion Pairs}}:
During practical inference, the generated motion can be aligned with the goal text at various levels (\eg, highly different, moderate, highly similar). Hence, it is necessary to train a model to evaluate and refine the initial motions with various alignment levels, and our idea is to sample negative text-motion pairs with various alignment levels to the anchor pair.
However, for a diverse text-motion dataset, simply random sampling negative pairs is not feasible, as there is an extremely high probability of selecting a text-motion that significantly differs from the anchor one. 
To address this, we design a ranking-based sampling strategy based on a pre-trained motion encoder \cite{tmr}, which is shown in the upper part of \cref{fig:date_engine}.

Suppose we attempt to sample $N_{neg}$ text-motion pairs $\{(t_i, m_i)\}_{i=1}^{N_{neg}}$ for an anchor pair $(t_{anc}, m_{anc})$. We first extract the motion embeddings for all motions in the dataset, compute the embedding distances across anchor motion $m_{anc}$ and all other motions in the dataset, and rank the distance from high to low (\ie, highly different to highly similar). By doing so, we obtain a ranked instance pool for the anchor motion. After that, for each time, we randomly sample a negative motion together with its text from the top $p$ fraction of highest-distance instances in the instance pool, and remove all this fraction of instances from the pool.

\vspace{0.1cm}
\noindent{\textbf{Text-Motion Alignment \& Refinement Instructions}}:
After assigning negative samples for each text-motion pair in the dataset, as shown in the lower part of \cref{fig:date_engine}, we further instruct LLMs to perform alignment assessment (including the text-motion \textbf{alignment evaluations} and \textbf{refinement instructions} for improvement) based on the text descriptions of anchor and negative samples. 

\vspace{0.1cm}
\noindent{\textbf{Annotation Quality Checking}}: After obtaining the annotations from LLM, we manually check the annotations and filter out the failure cases (e.g., empty responses, refinements to already correct motions) prior to model training. We discuss how the filtering strategy influences the performance in the Appendix.

\vspace{0.1cm}
\noindent{\textbf{Data Organization for Training}}: For Stage 1, based on the augmented annotations mentioned above, we can construct training data for all tasks introduced in \cref{sec:IR_init} with task-specific instructions. For Stage 2, we organize the CoT trajectories by concatenating the goal analysis, motions, assessment results, and pre-defined plans to match the CoT template. For the \emph{intermediate motions}, we sort the selected negative text-motion pairs in an alignment-improving rank, and insert them into the reasoning trajectory. For Stage 3, the data can be directly selected from existing motion-text-paired or text-only datasets without further organization. 
Please refer to the Appendix for more details.

\section{Experiments}
\vspace{-0.2cm}
\subsection{Setups}
\vspace{-0.1cm}
\noindent{\textbf{Dataset}}. Following previous works on MoLLM \cite{motiongpt, motionchain, mg-motionllm,motionagent}, we mainly conduct experiments on HumanML3D dataset \cite{guo}, a large-scale benchmark for motion generation and understanding. 
We also compare our approach with existing methods on KIT-ML \cite{kit-ml}, another popular Text-to-Motion dataset.
For Stage-1 and Stage-2 training, we build the instruction tuning dataset by using the data engine introduced in \cref{sec:data_engine}. For Stage-3 training, we sample approximately 9k data for each dataset. Specifically, for experiments on the HumanML3D, we source data from the original \cite{guo} and InstructMotion \cite{sheng2024exploring} dataset. 
Note that prompts in the InstructMotion dataset are also sourced from HumanML3D but augmented by LLM. 
For experiments on the KIT-ML dataset, we source data from the original KIT-ML dataset and use GPT-4o to augment the original prompts.
\emph{We strictly ensure there are no identical prompts in Stage-3 data with those in the testing set for fair evaluation.}
See Appendix for more details.

\vspace{0.1cm}
\noindent{\textbf{Evaluation Metrics}}.
For the text-to-Motion task, we follow previous works to report metrics including R-Precision, FID, MM-Dist, and Diversity. For most of the evaluation, we adopt the official evaluators from \cite{guo}. We also employ a new evaluator from MARDM \cite{mardm} for further evaluation. We conduct each evaluation 20 times, presenting the average metrics.
Besides, as our stage-1 model can also perform motion understanding tasks, we also conduct experiments on the Motion-to-Text caption task and adopt NLP metrics including Bleu\cite{bleu}, Rouge-L\cite{rouge}, CIDEr\cite{cider}, and BertScore\cite{bertscore} following previous works \cite{motiongpt, motionagent, motionchain}. See Appendix for more details about the metrics.

\vspace{0.1cm}
\noindent{\textbf{Implementation Details}}.
In experiments on the HumanML3D dataset, we initialize our base model with the official weights of MotionLLM \cite{motionagent}, which is pre-trained on text-to-motion generation task.
For experiments on the KIT-ML dataset, we train a base model based on the official implementation of \cite{motionagent} with text-to-motion generation task.
Please refer to the Appendix for more details.

\vspace{-0.1cm}
\subsection{Diving Deeper into IRG-MotionLLM}
\label{sec:main_ablations}
\vspace{-0.1cm}

\noindent{\textbf{Impacts of training tasks in IRMoGen-Initialization (Stage-1)}}. We ablate the training tasks and evaluate the stage-1 model on the text-to-motion generation task. As shown in \cref{tab:ablation_all_stages} Row1-4, further fine-tuning the base model with dense motion token supervision consistently brings improvement on FID. However, with only the T2M task (Row 1 v.s. Row 2), the model suffers from deterioration on all other metrics. We hypothesize it is because the base model is already converged on the T2M task and tends to overfit the training data when further trained on the same task instead of learning new text-motion alignment knowledge. 
Introducing other basic tasks (Row 1 v.s. Row 3) slightly benefits Top-1, but the limited improvement suggests that there exists a task gap limiting the complementary effects between learning basic motion understanding and generation tasks. By combining all tasks (Row 1 v.s. Row 4), consistent improvement on all metrics is observed. Not coincidentally, we also observe the same performance trend on the motion-to-text task (see \cref{tab:ablation_s1_tasks_m2t}).
These phenomena suggest that \textbf{\emph{Text-Motion Assessment and Refinement tasks are the key to connect basic motion generation and understanding tasks}}.

\vspace{0.1cm}
\noindent{\textbf{Impacts of IRMoGen-CoT Learning (Stage2)}}. As shown in \cref{tab:ablation_all_stages}, training the model on IRMoGen-CoT explicitly connected motion understanding and generation knowledge, resulting in higher alignment between generated and goal text with only one-round input (Row 4 v.s. Row 7). We also compare two other variants: (1) w/o Asse.+Ref.: Building Stage-2 training data with initial goal analysis and one-round generation without multi-round assessment and refinement. 
(2) w/o I.I.: Ablating the ``Ignore Incorrect'' strategy discussed in \cref{sec:IR_enhance}.
Comparisons between Row 5 and Row 7 further showcase the effectiveness of IRMoGen-CoT Learning. Comparison between Row 6 and Row 7 suggests that training the model to generate incorrect \emph{intermediate motions} will significantly break the already learned alignment between motion and text.

\begin{table}[t]
    \centering
    \begin{minipage}[t]{0.48\textwidth}
        \scriptsize
        \centering
        \scriptsize
        \caption{\textbf{Ablation studies of training stages on HumanML3D text-to-motion generation task.} *: We use the official weights of \cite{motionagent} as our base model.} 
        \vspace{-0.3cm}
        \resizebox{\linewidth}{!}{
            \begin{NiceTabular}{l|l|cccccc}
            \toprule
            \multirow{2}{*}{} & \multirow{2}{*}{{Methods}}  & \multicolumn{2}{c}{R-Precision$\uparrow$} & \multirow{2}{*}{FID$\downarrow$} &  \multirow{2}{*}{MM-Dist$\downarrow$} & \multirow{2}{*}{Diversity$\uparrow$}   \\ 
            \cline{3-4}
            &&Top-1 &  Top-3\\
            \midrule
            & \multicolumn{6}{l}{\emph{\textbf{Baseline Model}}} \\
            1 & \scriptsize MotionLLM* \cite{motionagent} \tiny {ICLR'25}  & 0.496  & 0.782 & 0.198 & 3.029 & 9.864\\
            \midrule
            & \multicolumn{6}{l}{\emph{\textbf{IRMoGen-Initialization (Stage-1)}}} \\
            2 &{ w/ T2M Task Only} & 0.495   & 0.778 & 0.185 & 3.072 & 9.734\\
            3 &{ w/ Basic Tasks Only} & 0.500  & 0.774 & 0.123 & 3.086 & 9.720\\
            4 & \rowcolor{lightpurple}Ours (S1)  & {0.504}  & {0.785} & {0.141} & {3.018} & {9.814}\\
            \midrule
            & \multicolumn{6}{l}{\emph{\textbf{IRMoGen-CoT Learning (Stage-2)}}} \\
            5 & w/o Asse.+Ref. & 0.503  & 0.782& 0.125 & 3.023 & 9.680 \\
            6 & w/o I.I.  & 0.448  & 0.720 & 0.592 & 3.519 & 9.675\\
            7 & \rowcolor{lightpurple}Ours (S2)  & {0.526}  & {0.810} & {0.111} & {2.885} & {9.819} \\
            
            \midrule
            & \multicolumn{6}{l}{\emph{\textbf{IRMoGen Reinforcing (Stage-3)}}} \\
            8 &\rowcolor{lightpurple}Ours (S3)  & {0.535}  & {0.820} & 0.242 & {2.785} &  9.900 \\
            \bottomrule   
            \end{NiceTabular}
        }
        \label{tab:ablation_all_stages}
    \end{minipage}
    \hfill  
    \begin{minipage}[t]{0.48\textwidth}
        \centering
        \scriptsize
        \vspace{1cm}
        \caption{\textbf{Ablation studies of training tasks on the HumanML3D motion caption task.} Combining both the basic and improving tasks brings stable improvement in text-motion alignment.} 
        \vspace{-0.3cm}
        \resizebox{\linewidth}{!}{
            \begin{NiceTabular}{l|ccccc}
            \toprule
            
            {{Methods}} & Bleu-1$\uparrow$& Bleu-4$\uparrow$& Rouge$\uparrow$ & CIDEr$\uparrow$ & BERT$\uparrow$\\
            \midrule 
            { w/ M2T Task Only} & 58.73 & 20.53 & 46.12 & 47.27 & 40.99\\
            { w/ Basic Tasks Only} & 59.77& 21.07 & 46.50 & 48.94 & 41.46\\
            \rowcolor{lightpurple}Ours (S1) & {62.05} & {22.53} & {47.51} & {52.4} & {42.68}\\
            
            \bottomrule   
            \end{NiceTabular}
        }
        \label{tab:ablation_s1_tasks_m2t}
    \end{minipage}
        \vspace{-0.3cm}
\end{table}

\begin{figure}[t]
    \begin{minipage}{0.48\textwidth}
        \centering
        \includegraphics[width=1\linewidth]{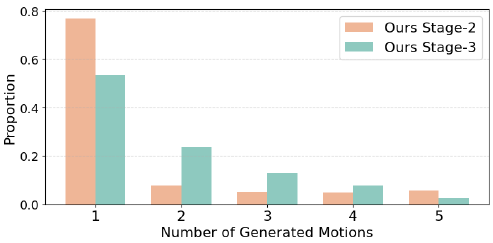}
        \vspace{-0.7cm}
        \caption{\textbf{Distributions of the number of generated motions in IRMoGen process}. Our method responses with longer reasoning with more intermediate generated motions after RL-tuning.}
        \label{fig:num_gens}
\end{minipage}
\hfill  
\begin{minipage}{0.48\textwidth}
        \centering
        \scriptsize
        \vspace{-0.5cm}
        \captionof{table}{\textbf{Ablation Studies on IRMoGen on HumanML3D dataset.} We observe clear improvement after adopting Interleaved Reasoning with multi-round motion assessment and refinement.} 
        \resizebox{\linewidth}{!}{
            \begin{NiceTabular}{l|cccccc}
            \toprule
            \multirow{2}{*}{{Methods}}  & \multicolumn{3}{c}{R-Precision$\uparrow$} & \multirow{2}{*}{FID$\downarrow$} &  \multirow{2}{*}{MM-Dist$\downarrow$} & \multirow{2}{*}{Diversity$\uparrow$}   \\ 
            \cline{2-4}
            &Top-1 & Top-2 & Top-3\\
            \midrule
            Ours (S1) init-gen &  0.504 & 0.693 & 0.785 & 0.141& 3.018 & 9.814\\
            Ours (S1) final-gen & 0.522 & 0.711 & 0.805 & 0.135	&2.906 & 9.740\\
            \midrule
            \scriptsize Ours (S2) init-gen  & 0.497 & 0.686 & 0.782 & 0.136 & 3.052 & 9.825\\
            \scriptsize Ours (S2) final-gen  & 0.526 & 0.717 & 0.810 & 0.111 & 	2.885 & 9.819 \\
            \midrule
            \scriptsize Ours (S3) init-gen  & 0.503 & 0.687 & 0.782 & 0.221 & 3.006 & 9.853 \\
            \scriptsize Ours (S3) final-gen  & {0.535} & {0.725} & {0.820} & 0.242 & {2.785} &  9.900\\ 
            \bottomrule   
            \end{NiceTabular}
        }
        \vspace{-0.5cm}
        \label{tab:ablation_multi-gen}
    \end{minipage}
    \vspace{-0.2cm}
\end{figure}

\vspace{0.1cm}
\noindent{\textbf{Impacts of IRMoGen Reinforcing (Stage3)}}. 
As shown in the last two rows in \cref{tab:ablation_all_stages}, with GRPO-based IRMoGen reinforcement learning, our model achieves further improvement on text-motion alignment. In alignment with the conclusion of previous work \cite{yuan2025ar}, vanilla GRPO can bring fluctuation to FID due to the lack of dense supervision on generated motion tokens. 
However, it is still competitive with those motion-aware LLMs \cite{motiongpt, motionchain,mg-motionllm} that support both text and motion outputs. 
User study in the Appendix and the supplementary videos further confirm the semantic alignment and naturalness of the generated motions.
Additionally, we visualize how IRMoGen Reinforcing influences the reasoning traces and text-motion alignment. As shown in \cref{fig:num_gens}, while our Stage-2 model learn to match the pattern of IRMoGen CoT, we observe that for most of the cases (more than 75\%), the  model tends to end the reasoning process after initial goal analysis and once generation without further refinement. After RL tuning, the model learns to extend the length of reasoning traces and perform multi-round motion assessment and refinement.

\vspace{0.1cm}
\noindent{\textbf{Initial Generation v.s. Final Generation}}. To evaluate the effectiveness of IRMoGen, we compare the initially and finally generated {motions} in parallel across all stages. Note that for the stage-1 model, we adopt the strategy discussed in \cref{sec:IR_init} to perform IRMoGen. 
From \cref{tab:ablation_multi-gen}, we have consistent observations: \textbf{\emph{Interweaving motion generation, assessment, and refinement significantly benefits the alignment between generated motions and goal texts}}.
More surprisingly, manually connecting relevant tasks on our Stage1 model can already bring clear improvement (\eg, 0.504 to 0.522 on Top-1), even if it is not explicitly trained to perform IRMoGen. 
We also discuss the impact of rounds of refinement in the Appendix.

\begin{table}[t]
    \scriptsize
    \centering
    \scriptsize
    \caption{\textbf{Compared with existing VQ-based Motion-aware LLMs on Text-to-Motion task.} *: We use the official weights of MotionLLM \cite{motionagent} as our base model on the HumanML3D dataset. $\S$: We train the base model following the official implementation of MotionLLM \cite{motionagent} on the KIT-ML dataset. Models with $\clubsuit$ support both text and motion outputs. We also report 95\% confidence intervals.} 
    \vspace{-0.3cm}
    \resizebox{1\linewidth}{!}{
        \begin{NiceTabular}{lcc|cccccc}
        \toprule
        
        \multirow{2}{*}{{Methods}} & \multirow{2}{*}{{Framework}} & \multirow{2}{*} {w/ GRPO} & \multicolumn{3}{c}{R-Precision$\uparrow$} & \multirow{2}{*}{FID$\downarrow$} &  \multirow{2}{*}{MM-Dist$\downarrow$} & \multirow{2}{*}{Diversity$\uparrow$}   \\ 
        \cline{4-6}
        &&&Top-1 & Top-2 & Top-3\\
        \midrule 
        \multicolumn{9}{c}{\emph{\textbf{\small HumanML3D Dataset}}} \\
        T2M-GPT\cite{t2m-gpt} \tiny {CVPR'23} \scriptsize & VQ+GPT2 & $\times$& $0.492^{\pm.003}$ & $0.679^{\pm.002}$ & $0.775^{\pm.002}$ & $0.141^{\pm.005}$ & $3.121^{\pm.009}$ & $9.722^{\pm.082}$ \\

        MotionGPT$\clubsuit$ \cite{motiongpt}\tiny NeurIPS'23 \scriptsize & VQ+T5 & $\times$ & $0.492^{\pm.003}$ & $0.681^{\pm.003}$ & $0.733^{\pm.006}$ & $0.232^{\pm.008}$ & $3.096^{\pm.008}$ & $9.528^{\pm.071}$ \\

        MotionChain$\clubsuit$ \cite{motionchain}\tiny ECCV'24 \scriptsize & VQ+T5 & $\times$ & $0.504^{\pm.003}$ & $0.617^{\pm.002}$ & $0.790^{\pm.003}$ & $0.248^{\pm.009}$ & $3.033^{\pm.010}$ & $9.470^{\pm.075}$ \\
        MG-MotionLLM$\clubsuit$\cite{mg-motionllm}\tiny CVPR'25 \scriptsize & VQ+T5 & $\times$& $0.516^{\pm.002}$ & $0.706^{\pm.002}$ & $0.802^{\pm.003}$ & $0.303 ^{\pm.010}$ & $2.952^{\pm.009}$ & $9.960^{\pm.073}$ \\
        ViMoRAG\cite{vimorag}  \tiny {NeurIPS'25} \scriptsize & VQ+Phi3-3.8B & $\times$& $0.452 ^{\pm.014}$ & $0.655^{\pm.014}$ & $0.764^{\pm.013}$ & $0.131^{\pm.073}$ & $3.146^{\pm.015}$ & $9.424^{\pm.001}$\\
        MotionGPT-2$\clubsuit$ \cite{motiongpt2}\tiny ArXiv'24 & VQ+LLaMA3.1-8B & $\times$ & $0.496^{\pm.002}$ & $0.691^{\pm.003}$ & $0.782^{\pm.004}$ & $0.191^{\pm.004}$ & $3.080^{\pm.013}$ & $9.860^{\pm.026}$ \\
        MotionGPT-2$\clubsuit$\cite{motiongpt2}\tiny ArXiv'24 & VQ+Gemma-2B & $\times$ & 0.436 & 0.600 & 0.697 & 0.228 & 3.589 & \textbf{10.081} \\
        \scriptsize MotionLLM \cite{motionagent}\tiny {ICLR'25} \scriptsize & VQ+Gemma2-2B & $\times$ & $0.515^{\pm.004}$ & $0.691^{\pm.003}$ & $0.801^{\pm.004}$ & $0.230^{\pm.009}$ & $2.967^{\pm.020}$ & $9.908^{\pm.102}$ \\
        \scriptsize MotionLLM* \cite{motionagent} \tiny {ICLR'25}  & VQ+Gemma2-2B & $\times$& ${0.496}^{\pm.002}$ & ${0.684}^{\pm.002}$ & ${0.782}^{\pm.002}$ & ${0.198}^{\pm.006}$ & ${3.029}^{\pm.007}$ & ${9.864}^{\pm.007}$\\
        \rowcolor{lightpurple}\scriptsize Ours (S1, w/ IRMoGen)$\clubsuit$   & VQ+Gemma2-2B & $\times$&$0.522^{\pm.003}$& $0.711^{\pm.002}$& $0.805^{\pm.002}$ & $0.135^{\pm.005}$	& $2.906^{\pm.009}$ & $9.740^{\pm.080}$\\
        \rowcolor{lightpurple}\scriptsize Ours (S2, w/ IRMoGen)$\clubsuit$ & VQ+Gemma2-2B & $\times$& $\textbf{0.526}^{\pm.002}$ & $\textbf{0.717}^{\pm.002}$ & $\textbf{0.810}^{\pm.002}$ & $\textbf{0.111}^{\pm.005}$ & $\textbf{2.885}^{\pm.008}$& $9.819^{\pm.088}$\\
        \midrule
        Motion-R1$\clubsuit$ \cite{motion-r1}\tiny ICLR'26 \scriptsize & VQ+Qwen2.5-3B & $\checkmark$ & $0.515^{\pm.003}$ & $0.719^{\pm.002}$ & $0.818^{\pm.002}$ & $\textbf{0.201}^{\pm.004}$ & $2.854^{\pm.010}$ & $\textbf{10.026}^{\pm.075}$ \\
        \rowcolor{lightpurple}\scriptsize Ours (S3, w/ IRMoGen)$\clubsuit$ & VQ+Gemma2-2B & $\checkmark$ & $\textbf{0.535}^{\pm.002}$ & $\textbf{0.725}^{\pm.002}$ & $\textbf{0.820}^{\pm.002}$ & ${0.242}^{\pm.006}$ & $\textbf{2.785}^{\pm.006}$ &  ${9.900}^{\pm.094}$\\
        \midrule
        \multicolumn{9}{c}{\emph{\textbf{\small KIT-ML Dataset}}} \\
        T2M-GPT\cite{t2m-gpt} \tiny {CVPR'23} \scriptsize & VQ+GPT2 &  $\times$& $0.416^{\pm.006}$ & $0.627^{\pm.006}$ & $0.745^{\pm.006}$ & $0.514^{\pm.029}$ & $3.007^{\pm.023}$ & $10.921^{\pm.108}$ \\
        MotionGPT$\clubsuit$ \cite{motiongpt}\tiny NeurIPS'23 \scriptsize & VQ+T5 & $\times$ & $0.366^{\pm.005}$ & $0.558^{\pm.004}$ & $0.680^{\pm.005}$ & $0.510^{\pm.016}$ & $3.527^{\pm.021}$ & $10.350^{\pm.084}$\\
        MotionGPT-2$\clubsuit$ \cite{motiongpt2}\tiny ArXiv'24 & VQ+Gemma-2B & $\times$ & $0.364$ & $0.581$ & $0.699$ & $1.063$ & $3.424$ & $10.603$\\
        \scriptsize MotionLLM$\S$ \cite{motionagent} \tiny {ICLR'25} & VQ+Gemma2-2B &  $\times$& $0.416^{\pm.004}$ & $0.637^{\pm.004}$ & $0.746^{\pm.003}$ & $0.560^{\pm.025}$ & $3.001^{\pm.011}$ & $\textbf{11.209}^{\pm.135}$\\
        \rowcolor{lightpurple}\scriptsize Ours (S1, w/ IRMoGen)$\clubsuit$  & VQ+Gemma2-2B &  $\times$& ${\textbf{0.425}^{\pm.005}}$ & $0.640^{\pm.004}$ & $0.754^{\pm.002}$ & $0.497^{\pm.021}$ & $\textbf{2.877}^{\pm.010}$ & $11.114^{\pm.095}$\\
        \rowcolor{lightpurple}\scriptsize Ours (S2, w/ IRMoGen)$\clubsuit$ & VQ+Gemma2-2B &  $\times$&  $0.419^{\pm.006}$ & $\textbf{0.642}^{\pm.006}$ & $\textbf{0.763}^{\pm.005}$ & $\textbf{0.389}^{\pm.014}$ & $2.908^{\pm.026}$ & $11.096^{\pm.121}$ \\
        \midrule
        Motion-R1$\clubsuit$ \cite{motion-r1}\tiny ICLR'26 \scriptsize & VQ+Qwen2.5-3B &  $\checkmark$ & $0.431^{\pm.003}$ & $0.638^{\pm.002}$ & $0.761^{\pm.003}$ & $\textbf{0.287}^{\pm.004}$ & $3.196^{\pm.040}$ & $10.875^{\pm.052}$ \\
        \rowcolor{lightpurple}\scriptsize Ours (S3, w/ IRMoGen)$\clubsuit$ & VQ+Gemma2-2B &  $\checkmark$ & $\textbf{0.445}^{\pm.005}$ & $\textbf{0.681}^{\pm.003}$ & $\textbf{0.781}^{\pm.004}$ & ${0.432}^{\pm.013}$ & $\textbf{2.740}^{\pm{.017}}$ & $\textbf{11.115}^{\pm .086}$\\
        \bottomrule   
        \end{NiceTabular}
    }
    \vspace{-0.2cm}
    \label{tab:comparison_t2m}
\end{table}

\begin{table}[t]
    \scriptsize
    \centering
    \scriptsize
    \caption{\textbf{More comparisons with GRPO-based methods on HumanML3D dataset.} $\ddagger$: We scale up the training data in Stage-3 to the similar scale as UniMo \cite{wang2026unimo} and extend the GRPO duration. Motion-R1 \cite{motion-r1} does not clearly state the size of RL data and training duration, but likely use the full training data of HumanML3D as UniMo \cite{wang2026unimo}. Scaling-up training data and extend the training duration significantly improve the performance on most of the metrics.} 
    \vspace{-0.3cm}
    \resizebox{0.8\linewidth}{!}{
        \begin{NiceTabular}{l|c|ccccccc}
        \toprule
        \multirow{2}{*}{{Methods}} &  {GRPO} & \multicolumn{3}{c}{R-Precision$\uparrow$} & \multirow{2}{*}{FID$\downarrow$} &  \multirow{2}{*}{MM-Dist$\downarrow$} & \multirow{2}{*}{Diversity$\uparrow$}   \\ 
        \cline{3-5}
        & Steps &Top-1 & Top-2 & Top-3\\
        \midrule 
        Motion-R1 \cite{motion-r1}\tiny ICLR'26 \scriptsize & - & $0.515$ & $0.719$ & $0.818$ & ${0.201}$ & $2.854$ & ${10.026}$ \\
        UniMo \cite{wang2026unimo}\tiny AAAI'26 \scriptsize & 14,000 & ${0.539}$ & ${0.738}$ & ${0.831}$ & ${0.177}$ & ${2.768}$ &  ${10.042}$\\
        \rowcolor{lightpurple}\scriptsize Ours &  900 & ${0.535}$ & ${0.725}$ & ${0.820}$ & ${0.242}$ & ${2.785}$ &  ${9.900}$\\
        \rowcolor{lightpurple}\scriptsize Ours$\ddagger$ &  2,600 & ${0.564}$ & ${0.754}$ & ${0.841}$ & ${0.208}$ & ${2.628}$ &  ${9.883}$\\
        \bottomrule   
        \end{NiceTabular}
    }
    \label{tab:more_comparison_grpo_model}
\end{table}

\begin{table}[t]
    \centering
    \begin{minipage}[t]{0.48\textwidth}
        \centering
        \scriptsize
        \vspace{-0.1 cm}
        \caption{\textbf{Compared with existing Text-to-Motion methods with a recently proposed evaluator \cite{mardm} on the HumanML3D dataset.} *: We use the official weights of MotionLLM \cite{motionagent} as our base model. Models with $\clubsuit$ support both text and motion outputs.} 
        \vspace{-0.3cm}
        \resizebox{\linewidth}{!}{
            \begin{NiceTabular}{l|ccccc}
            \toprule
            
            \multirow{2}{*}{{Methods}}  & \multicolumn{3}{c}{R-Precision$\uparrow$} & \multirow{2}{*}{FID$\downarrow$} &  \multirow{2}{*}{MM-Dist$\downarrow$}   \\ 
             \cline{2-4}
            &Top-1 & Top-2 & Top-3\\
            \midrule 
            \multicolumn{5}{l}{\emph{\textbf{Mask-Modeling-based Models}}} \\
            MMM\cite{mmm}\tiny{CVPR'24} & 0.487 & 0.683 & 0.782 & 0.132 & 3.359 \\
            MoMask\cite{momask}\tiny{CVPR'24} & 0.490 & 0.687 & 0.786 & 0.116 & 3.353\\
            MoMask+DisCoRD\cite{cho2025discord}\tiny{ICCV'25} & 0.504 & 0.697 & 0.797 & \textbf{0.056} & 3.273\\
            \midrule 
            \multicolumn{5}{l}{\emph{\textbf{Diffusion-based Models}}} \\
            MDM\cite{mdm}\tiny{ICLR'23} & 0.440 & 0.636 & 0.742 & 0.518 & 3.640 \\
            MotionDiffuse\cite{motiondiffuse}\tiny{TPAMI'24} & 0.450 & 0.641 & 0.753 & 0.778 & 3.490\\
            MLD\cite{mld}\tiny{CVPR'23} & 0.461 & 0.651 & 0.750 & 0.431 & 3.455\\
            ReMoDiffuse\cite{remodiffuse}\tiny{ICCV'23} & 0.468& 0.653 & 0.754 & 0.883 & 3.414\\
            Light-T2M\cite{light-t2m}\tiny{AAAI'25} & 0.481 & 0.676 & 0.774 & 0.072 & 3.281\\
            MARDM-SiT\cite{mardm}\tiny{CVPR'25} & $0.500$ & 0.695 & 0.795 & {0.114} & 3.270\\
            \midrule 
            \multicolumn{5}{l}{\emph{\textbf{LLM-based Models}}} \\ 
            T2M-GPT\cite{t2m-gpt}\tiny {CVPR'23} \scriptsize & 0.487 & 0.659 & 0.758 & 0.335 & 3.505\\
            MotionLLM*\cite{motionagent}\tiny{ICLR'25} & 0.470 & 0.664 & 0.771 & 0.240 & 3.377\\
            \rowcolor{lightpurple} Ours (S1, w/ IRMoGen) $\clubsuit$ & 0.507 & 0.696 & 0.791 & 0.161 & 3.234 \\
            \rowcolor{lightpurple} Ours (S2, w/ IRMoGen) $\clubsuit$ & 0.503 & 0.698 & 0.796 & 0.185 & 3.212 \\
            \rowcolor{lightpurple} Ours (S3, w/ IRMoGen) $\clubsuit$ & \textbf{0.509} & \textbf{0.701} & \textbf{0.802} & 0.259 & \textbf{3.166} \\
            \bottomrule   
            \end{NiceTabular}
        }
        \vspace{-0.3cm}
        \label{tab:comparison_t2m_mardm_eval}
    \end{minipage}
    \hfill  
    \begin{minipage}[t]{0.48\textwidth}
        \centering
        \scriptsize
        \vspace{-0.1cm}
        \caption{\textbf{Compared with existing methods on Motion-to-Text Caption task on the HumanML3D dataset.} We strictly follow the evaluation protocol as discussed in \cite{motionagent, motiongpt} {Our Stage-1 model outperforms other methods on most of the metrics.}}
        \vspace{-0.3cm}
        \resizebox{\linewidth}{!}{
            \begin{NiceTabular}{l|ccccc}
            \toprule
            {{Methods}} & Bleu-1$\uparrow$& Bleu-4$\uparrow$& Rouge$\uparrow$ & CIDEr$\uparrow$ & BERT$\uparrow$\\
            \midrule 
            \multicolumn{5}{l}{\emph{\textbf{w/ Task Specific Tuning}}} \\
            TM2T\cite{tm2t} & 48.90 & 8.27 & 38.10 & 15.80 & 32.20 \\
            MG-MotionLLM\cite{mg-motionllm} & - & 8.06 & - & - & 36.70 \\
            LaMP\cite{lamp} &  47.80 & 13.04 & 37.10 & 28.90 & 32.70 \\
            MotionLLM\cite{motionagent} &  54.53 & 17.65 &\textbf{48.70} & 33.74 & {42.63}\\
            MoTe\cite{mote} & 46.70 & 11.15 & 37.40 & 31.50 & 30.30\\
            MotionGPT3\cite{motiongpt3} & 56.36 & 17.66 & 45.00 & 30.98 & 35.85\\
            \midrule 
            \midrule 
            \multicolumn{5}{l}{\emph{\textbf{w/o Task Specific Tuning}}} \\       
            MotionGPT \cite{motiongpt}& 48.2 &12.47 & 37.4 & 29.2 & 32.4\\
            MotionChain \cite{motionchain}& 48.1 & 12.56 & 33.9 & 33.7 & 36.9\\
            MG-MotionLLM \cite{mg-motionllm} & - & - & - & - & 38.6\\
            MotionGPT3\cite{motiongpt3} & 59.08 & 19.41 & 46.17 & 28.72 & 35.23\\
            Being-M0.5\cite{being_m05} & - & - & - & 26.78 & - \\
            \rowcolor{lightpurple}  Ours (S1)& \textbf{62.05} & \textbf{22.53} &{ 47.51} & \textbf{52.4} & \textbf{42.68}\\
            \bottomrule   
            \end{NiceTabular}
        }
        \vspace{-0.4cm}
        \label{tab:comparison_m2t}
    \end{minipage}
\end{table}

\vspace{0.1cm}
\noindent\textbf{Robustness evaluation on IRG-MotionLLM}. In Appendix, we show that our method goes beyond relying on specific reasoning traces but performs robust IRMoGen even if we introduce perturbations to the native reasoning traces.

\vspace{0.1cm}
\noindent\textbf{Emergent cross-model and cross-task synergies.} 
In the Appendix, we show that IRG-MotionLLM can be functioned as a text-motion reward model to enhance existing motion generator. On AToM-general benchmark \cite{atom}, our method outperforms other advanced reward models \cite{achiam2023gpt,motionpatch}, showcasing the quality and transferability of its motion assessment ability.
Moreover, evaluation on MotionFix benchmark \cite{motionfix} shows that IRG-MotionLLM also benefits the adaptation to the motion editing task and outperforms the baseline models \cite{motionagent,motionfix}.

\vspace{-0.1cm}
\subsection{IRG-MotionLLM v.s. Existing Methods}
\vspace{-0.1cm}
\noindent{\textbf{Comparisons with VQ-based Motion-aware LLMs}}.
In \cref{{tab:comparison_t2m}}, we compare our IRG-MotionLLM with previous methods with a similar framework on the HumanML3D and KIT-ML datasets. \textbf{First}, compared with our base model MotionLLM \cite{motionagent}, our training scheme brings stable improvement. \textbf{Second}, our Stage-1 and Stage-2 models outperform GRPO-free methods on most of the metrics. \textbf{Third}, equipped with GRPO-based RL, our Stage-3 model achieves further improvement on text-motion alignment and outperforms Motion-R1 \cite{motion-r1}. 
In alignment with the conclusion of previous work \cite{yuan2025ar}, vanilla GRPO can bring fluctuation to FID due to the lack of dense supervision on output motion tokens. However, it is still competitive with methods \cite{motiongpt,motionchain,mg-motionllm,motiongpt2,motion-r1} that support both text and motion outputs.
User studies in the Appendix further confirm the semantic alignment and naturalness of the generated motions.

\noindent{\textbf{More Discussions on Stage-3}}. By default, we use 9k data and train our model for 900 steps in the experiments on HumanML3D dataset. 
However, for the concurrent GRPO-based Motion-aware LLMs, the GRPO training stage is conducted with much larger motion-text pairs and lasts for longer steps (\eg, UniMo\cite{wang2026unimo} uses more than 23k motion data and optimizes the model for 14,000 GRPO steps).
Hence, a natural exploration is to scale up the training data and extend the training duration for our IRG-MotionLLM. We scale up the training data in Stage-3 to the similar scale as UniMo \cite{wang2026unimo} and train our model for 2,600 GRPO steps, noted as ``\emph{Ours$\ddagger$}''. As shown in \cref{tab:more_comparison_grpo_model}, scaling-up training data and extending the training duration significantly performance on most of the metrics (R-Precision, FID, and MM-Dist). 
By aligning the training setting, our method achieves a more competetive FID with existing GRPO-based Motion-aware LLMs, but shows clear superiority on motion-text alignment metrics.

\begin{figure}[t]
  \centering
   \includegraphics[width=1\linewidth]{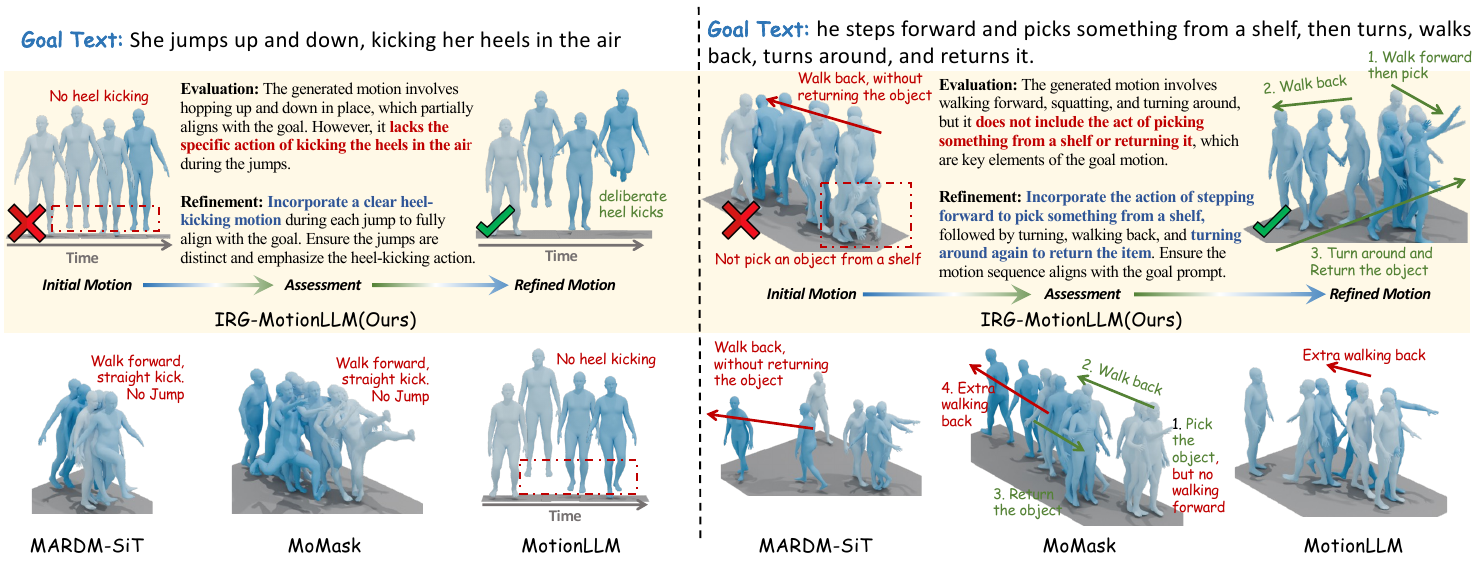}
    \vspace{-0.7cm}
   \caption{\textbf{Visualization results.}Native interleaved reasoning enables IRG-MotionLLM to: (1) recognize the misalignment between previously generated motion and the goal text and provide proper refinement instructions; (2) refine the motion based on previous reasoning. Such characteristics help our method more accurately follow the details of the goal text than existing methods. The human figures are colored from light to dark to indicate the progression of time. Zoom in for the best view.}
   \label{fig:case_study1}
   \vspace{-0.2cm}
\end{figure}

\vspace{0.1cm}
\noindent{\textbf{Comparisons with another Evaluator}}.
We further evaluate our method with a recently proposed evaluator \cite{mardm}, which is more robust to the redundant factors in motion representation. As shown in \cref{tab:comparison_t2m_mardm_eval}, a similar improvement trend is observed. Our model achieves advanced performance compared with existing methods built on various frameworks. 

\vspace{0.1cm}
\noindent\textbf{Comparisons on Motion Understanding Task}.
As shown in \cref{tab:comparison_m2t}, our Stage-1 model also achieves advanced performance in the Motion-to-Text Caption task without task-specific finetuning.

\vspace{0.1cm}
\noindent\textbf{More Comparisons}. In the Appendix, we further provide detailed analysis on the comparisons with more models with various frameworks \cite{lamp,hong2025salad,mogenst,bamm,being_m05,being-m0}.

\vspace{-0.1cm}
\subsection{Qualitative Results}
\vspace{-0.1cm}
In \cref{fig:case_study1}, we provide qualitative results of our method alongside SoTA motion generation methods. IRG-MotionLLM is able to natively perform interleaved reasoning across motion and text to improve the generation result. More visualizations (including the rendered videos) are provided in the Supplementary. 

\vspace{-0.1cm}
\section{Conclusions}
\vspace{-0.2cm}
We believe that beyond simply training models on various independent tasks, leveraging motion understanding to enhance generation holds vast research potential. To facilitate future exploration in this direction, we introduce Interleaved Reasoning for Motion Generation (IRMoGen), a novel paradigm that enhances final motion quality by seamlessly interleaving motion generation, assessment, and refinement within a unified reasoning process. We propose a three-stage training scheme to endow and further enhance the model’s capacity for native interleaved reasoning.
Extensive experiments evaluate the effectiveness and showcase the properties of our IRG-MotionLLM.
In Appendix, we further discuss the limitations of our work and outline future research directions. We expect our exploration and findings to offer fresh perspectives and inspire new advances in motion generation and motion-aware LLMs.

\section*{Acknowledgements}
This work was supported partially by NSFC(92470202), Guangdong NSF Project (No. 2023B1515040025), Guangdong Key Research and Development Program(No. 2024B0101040004, No.  2025B0909020002).
The authors thank Bowen Yin, Boyuan Sun, Zhi Ouyang and Wei-Jin Huang for the valuable discussions. 
The authors also thank anonymous reviewers and ACs for their constructive suggestions.

%
%

%
%
\bibliographystyle{splncs04}
\bibliography{main}

\clearpage
\newpage

\appendix
\section{Appendix Introduction}
This appendix includes more details and analysis of our works:

\vspace{0.2cm}
\noindent{\textbf{In \cref{sec:preliminaries}}}, we provide more preliminaries about: (1) Text-to-Motion Generation task definition; (2) Motion VQVAE; (3) Motion-aware LLM; (4) GRPO-based Reinforcement Learning.

\vspace{0.2cm}
\noindent\textbf{In \cref{sec:more_model_details}}, we provide more details of our IRG-MotionLLM, including: (1) Base Model; (2) Inference with IRMoGen.

\vspace{0.2cm}
\noindent\textbf{In \cref{sec:more_about_data_engine}}, we provide more details and discussions of the data engine, including: (1) More about the annotation pipeline; (2) More about training data organization; (3) Compare our data engine with the construction of MotionFix \cite{motionfix}.

\vspace{0.2cm}
\noindent\textbf{In \cref{sec:more_experiment_settings}}, we provide more details about the experiment setups, including: (1) Dataset; (2) Evaluation Metrics; (3) Implementation.

\vspace{0.2cm}
\noindent{\textbf{In \cref{sec:more_experiments}}}, we provide more experimental results and analysis, including:
    \begin{itemize}
        \item {Impact of the Manual Check on LLM-based annotations. (\cref{sec:ablation_manual_check})}
        \item {More Comparisons and Discussions on Text-to-Motion Generation Task. (\cref{sec:more_comparison_t2m})}
        \item {Unified Architecture v.s. Multi-Agent Pipeline. (\cref{sec:unified_vs_agentic})}
        \item {Impact of Multiple Round of Refinement. (\cref{sec:ablation_multi_round_refinement})}
        \item {Robust IRMoGen or Trajectory Dependency? (\cref{sec:discussion_rebustness})}
        \item {Stage-3 Training: Unleashing the capacity or Hacking the evaluator? (\cref{sec:discussion_grpo_hacking})}
        \item {Functioning IRG-MotionLLM as a Text-Motion Reward Model. (\cref{sec:irg_motionllm_as_reward})}
        \item {Adapting IRG-MotionLLM to Motion Editing task. (\cref{sec:irg_motionllm_to_editing})}
        \item {User studies. (\cref{sec:user_study})}
        \item {More Qualitative Results. (\cref{sec:more_visualizations})}
    \end{itemize}

\vspace{0.2cm}
\noindent\textbf{In \cref{sec:limitations}}, we discuss the limitations and the potential research directions, including: (1) IRMoGen with advanced MoLLM framework; (2) Motion Assessment beyond text-motion alignment; (3) Scaling-up IRG-MotionLLM; (4) Interleaved Reasoning for broader motion-centric tasks.

\section{Preliminaries}
\label{sec:preliminaries}
\subsection{Text-to-Motion Generation Task Definition}
Text-to-Motion Generation aims at generating 3D human motion aligned with the goal text. In this work, we follow \cite{guo,mdm} to represent the human motion sequence as  $m_{1:T}\in \mathbb{R}^{T\times D}$, where $T$ is the length (number of frames) of the motion sequence. Each human pose $m_i$ in the human motion sequence is a $D$-dimension vector, which includes the root angular velocity along the Y-axis, root linear velocities on the XZ-plane, root height, and local joint positions, velocities, and rotations relative to the root space.
Please refer to \cite{guo} for more detailed information on human pose representation.

\subsection{Motion VQVAE}
To align better with LLM’s next-token prediction mechanism, a Motion VQ-VAE \cite{motiongpt,mg-motionllm, motionagent} is adopted to encode motions into discrete representations and also decode these representations back to continuous motions. 

\vspace{0.2cm}
\textbf{Quantization}: Given a motion sequence $m_{1:T}\in \mathbb{R}^{T\times D}$, an encoder $\mathcal{E}$ first encodes it into a sequence of latent embeddings $z_{1:T/S}\in\mathbb{R}^{T/S\times d}$, where $S$ denotes the downsampling rate on temporal dimension and $d$ is the hidden dimensions. After that, the latent embeddings are quantized by a quantizer with a codebook $C=\{c_k\}^K_1$, which can be represented as: 

\vspace{-0.2cm}
$$
\hat{{z}}_t = \arg\min_{{C}_k \in {C}} \|{Z}_t - {C}_k\|_2,
$$
where $c_k\in\mathbb{R^d}$, $K$ is the size of the codebook, and $\hat{z}_{1:T/S}$ are the quantization results. 

\vspace{0.2cm}
\textbf{De-quantization}: Given the quantization results, a decoder $\mathcal{D}$ can de-quantize them back to the motion sequence: $\hat{m}_{1:T}=\mathcal{D}(\hat{z}_{1:T/S})$

\vspace{0.2cm}
\textbf{Training}: We share the same strategy as previous works \cite{motionagent} to optimize the VQVAE. The training loss is written as:
$$
L_{\text{VQVAE}} = ||m-\hat m||_1 + \alpha ||p-\hat p||_1 + \beta ||z-sg[\hat z]||_2.
$$
The first item is the motion reconstruction loss, the second item is a regularization on the joint positions $p$, the third item is the commitment loss. $sg[.]$ denotes the stop-gradient operation, $\alpha$ and $\beta$ are the hyper-parameters.
After training, we keep the Motion VQVAE frozen for further usage. 

\subsection{Motion-aware LLM}
Based on the Motion VQVAE, the motion sequence can be discretized into \emph{K} distinct motion tokens. We map these motion tokens to the motion token vocabulary $V_m=\{\textless Motion\_i\textgreater\}^K_{i=1}$ based on their indices. After that, we extend the base LLM vocabulary with these motion-specific codes alone with boundary tokens $\textless Motion\textgreater$ and $\textless /Motion\textgreater$ to delineate motion sequence spans within the text (\ie, K+2 new tokens).
After that, given prompt $q$ and target token sequence $x_{1:T}$, we can train a motion-aware LLM with standard next-token prediction objective via negative log-likelihood:
$$
    L_{LLM} = -\sum log{\space}p_\theta(x_t|x_{x<t},q).
$$
During inference, the model is able to freely mixed token sequence (with both text and motion tokens). We can directly extract the generated motion by matching the boundary tokens $\textless Motion\textgreater$ and $\textless /Motion\textgreater$.

\subsection{GRPO-based Reinforcement Tuning}
GRPO is initially proposed to enhance the reasoning capacity of LLM \cite{deepseek-math,guo2025deepseek}, and proven to have promising potential for enhancing Multi-modal LLM \cite{t2i-r1,vision-r1,motion-r1}.
The core mechanism of GRPO \cite{deepseek-math} is to leverages intra-group ranking feedback to optimize the policy (\ie, our IRG-MotionLLM) by encouraging better samples within each group, which eliminates the need of an additional critic model \cite{ppo,motionrl} or pair-wise preference data \cite{dpo,atom,pappa2024modipo}.

Specifically,  taken the reasoning instruction $q$, a group of $G$ trajectories $\{o_1,o_2,...,o_G\}$ are sampled from the model. After that, each trajectory is assigned with a scalar reward $r(o_i)$. 
GRPO then standardizes these scores within the group and optimizes a weighted objective:
\begin{equation}
    R(o) = \sum_{i=1}^{G} \frac{\pi_{\theta}(o_i|q)}{\pi_{b_{\text{old}}}(o_i|q)} \cdot 
\frac{r(o_i) - \operatorname{mean}\left(\{r(o_i)\}_{i=1}^{G}\right)}{\operatorname{std}\left(\{r(o_i)\}_{i=1}^{G}\right)},
\end{equation}
where $\pi_\theta$ indicates the current policy and $\pi_{\theta_{old}}$ is the previous policy. Based on $R(o)$, the objective is written as:
\begin{equation}
\max_{\pi_\theta} \mathbb{E}_{o \sim \pi_{\theta_{\text{old}}}(p)} \left[ R(o) - \beta_{kl} D_{\text{KL}}(\pi_\theta \| \pi_{\text{ref}}) \right],
\end{equation}
where $\pi_{ref}$ is indicates initial policy and $\beta_{kl}$ is a hyper-parameter.

\section{More Details about IRG-MotionLLM}
\label{sec:more_model_details}
\subsection{Base Model}
In our work, we mainly use the MotionLLM from the T2M part of MotionAgent \cite{motionagent} as our base model. The size of codebook is 512, and the LLM part is build on Gemma2-2B \cite{gemma2}.
Based on the training strategy introduced in \cref{sec:preliminaries}, the base model is trained on text-to-motion generation task. During training, only the LoRA adapter is trainable while the other parameters are frozen. Please refer to the original paper \cite{motionagent} for more details.

\begin{figure}[t]
  \centering
   \includegraphics[width=\linewidth]{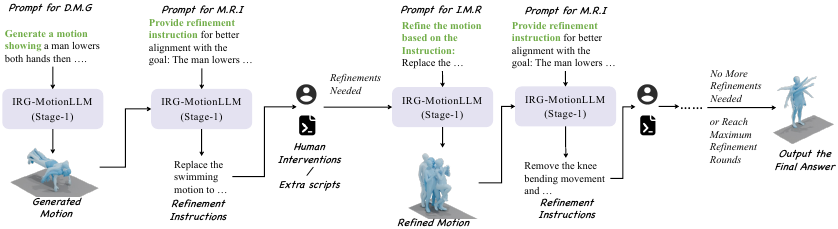}
    \vspace{-0.4cm}
   \caption{\textbf{The inference pipeline for performing IRMoGen on our Stage-1 model.} Our Stage-1 model can perform interleaved reasoning across generation, assessment and refinement tasks by sequentially using prompts of different tasks together with the outputs from previous steps. To determine when to stop, extra scripts or manual interventions are needed to check the refinement instruction. D.M.G: Direct Motion Generation. M.R.I: Motion Refinement Instructing. I.M.R: Instruction-guided Motion Refinement.}
   \label{fig:stage1_IRMoGen}
   \vspace{-0.2cm}
\end{figure}

\subsection{More details about Inference with IRMoGen}
\label{sec:inference_IRMoGen}
Prompted with reasoning instruction and goal text, our Stage-2 and Stage-3 models can natively perform IRMoGen, and adaptively stop reasoning when a satisfactory motion is obtained (\ie, the assessment results indicate ``\emph{no refinement is needed}'').
Notably, our Stage-1 model can also perform interleaved generation, assessment and refinement by sequentially using prompts for different tasks together with the outputs from previous steps. To know when to stop reasoning, extra scripts or manual interventions are needed to check the refinement instruction. The this inference pipeline is shown in \cref{fig:stage1_IRMoGen}.

Additionally, to avoid endless reasoning, we introduce the maximum refinement constraint $N_{max}$, \ie, the model can perform at most $N_{max}$ rounds of refinement. If the rounds of refinement reach $N_{max}$, the later reasoning will be pruned and use the last generated motion as the output.

\section{More Details about Data Engine}
\label{sec:more_about_data_engine}
\subsection{Details of Annotation Pipeline}
\label{sec:annot_pipeline_appendix}
\begin{figure}[t]
  \centering
   \includegraphics[width=0.99\linewidth]{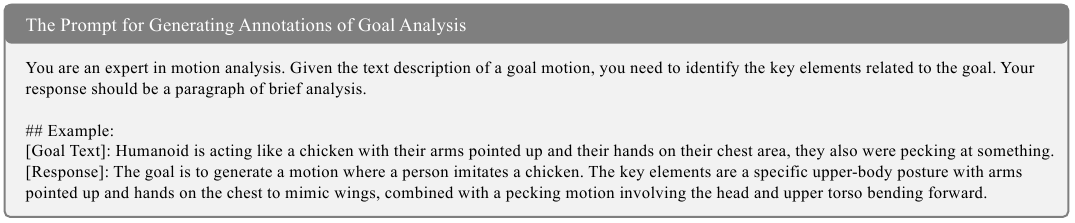}
    \vspace{-0.4cm}
   \caption{The prompts for generating annotations of Goal Analysis.}
   \label{fig:prompt_goal_analysis}
   \vspace{-0.2cm}
\end{figure}

\begin{figure}[t]
  \centering
   \includegraphics[width=0.99\linewidth]{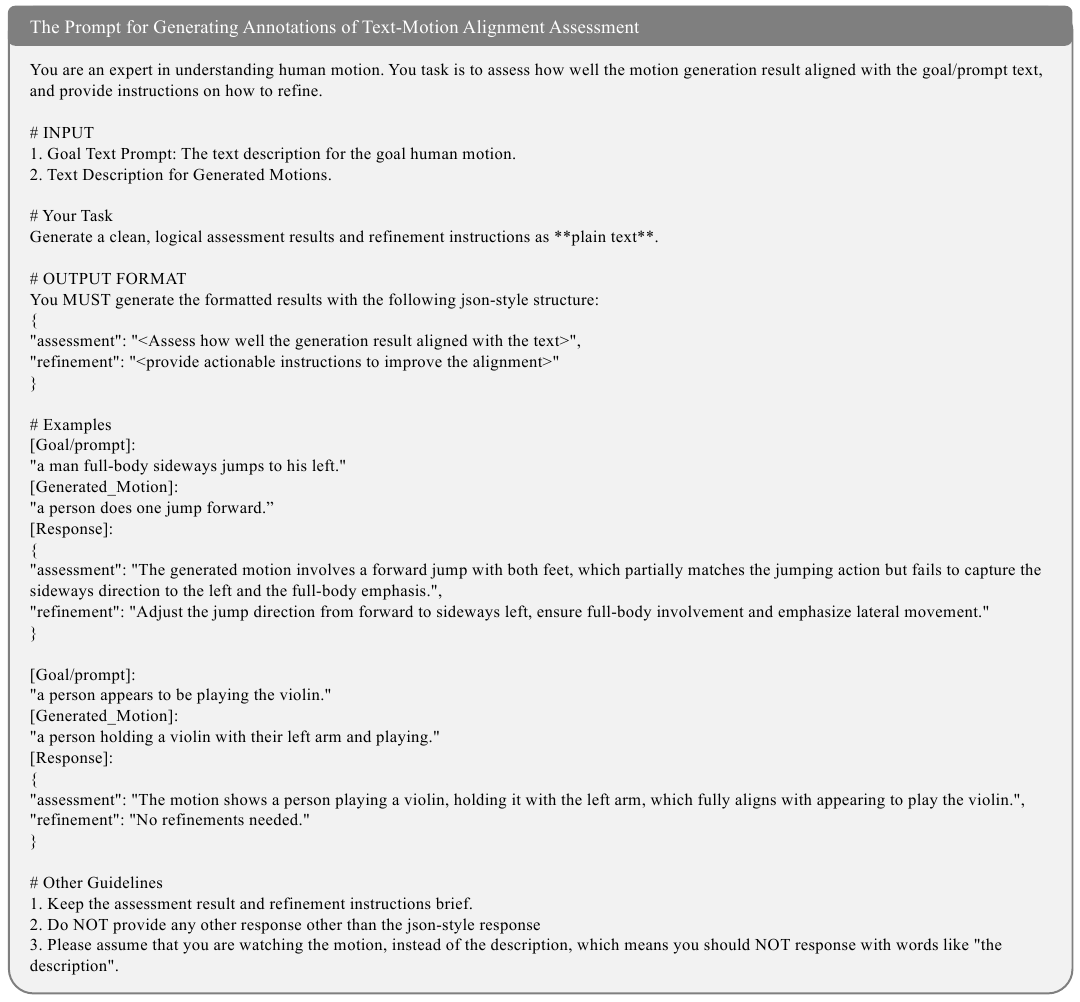}
    \vspace{-0.4cm}
   \caption{The prompts for generating annotations of Text-Motion Alignment Assessment.}
   \label{fig:prompt_tm_assessment}
   \vspace{-0.2cm}
\end{figure}

As discussed in Sec. \textcolor{red}{4}, for each anchor text-motion pair, we progressively sample $N_{neg}$ negative text-motion pairs from the ranked instance pool with a portion $p$. In our work, $N_{neg}$ is set to 4, and $p$ is set to 90\% (9/10) for the HumanML3D dataset and 87.5\% (7/8) for the KIT-ML dataset. 
Additionally, we prompt GPT-4o to obtain the goal analysis and the text-motion alignment assessment annotations. The detailed prompts are shown in \cref{fig:prompt_goal_analysis} and \cref{fig:prompt_tm_assessment}.

After collecting the LLM-generated annotations, we further conduct a double-check on the generated annotations. We filter out the invalid annotations (\eg, empty response or refining an already perfect motion) prior to model training. We further discuss the impact of the filtering strategy in \cref{sec:ablation_manual_check}.

\subsection{Training data organization for Stage-1}
\begin{figure}[t]
  \centering
   \includegraphics[width=1\linewidth]{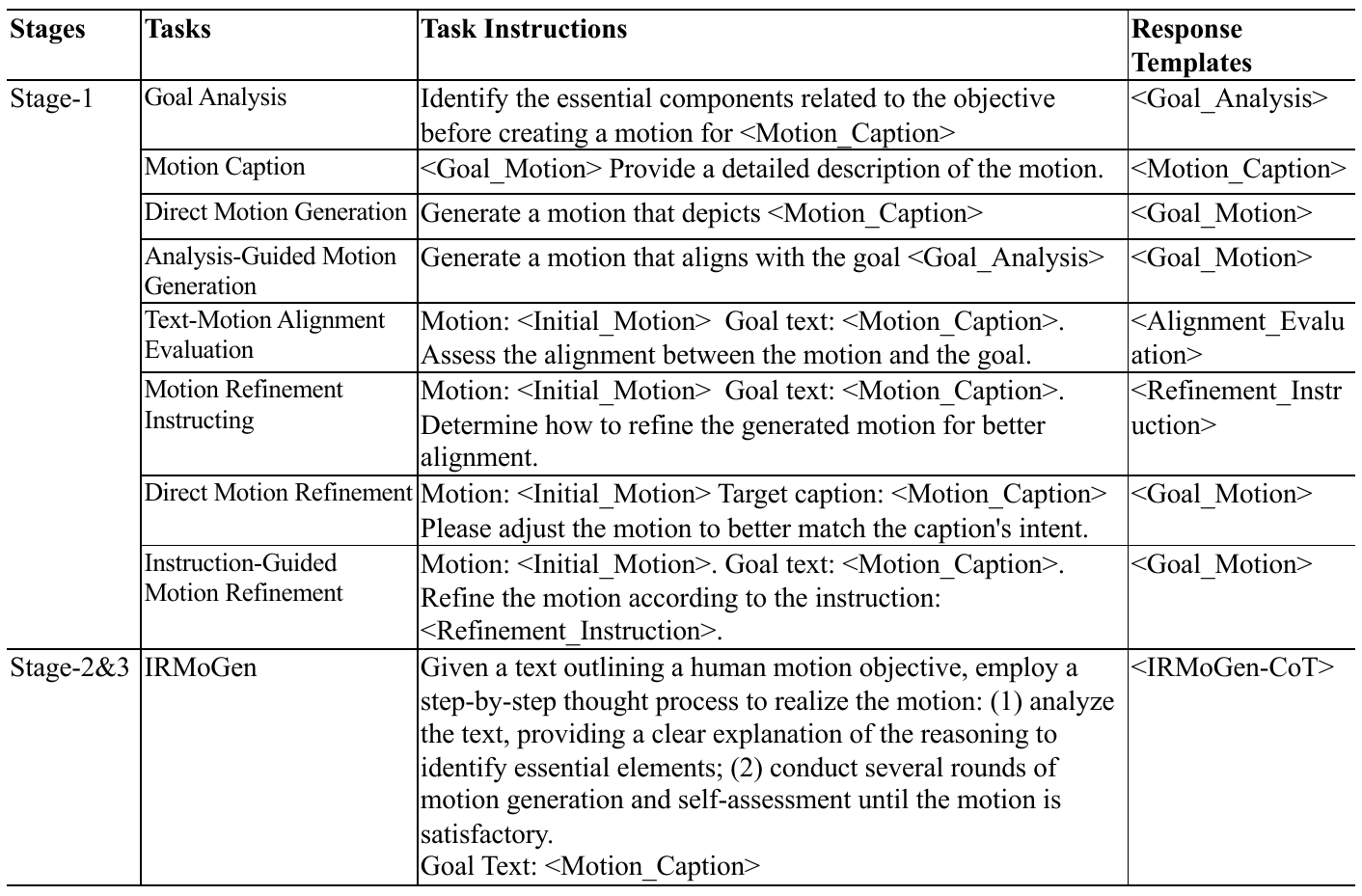}
    \vspace{-0.4cm}
   \caption{\textbf{The task instructions and response templates.} The fields between ``\textless'' and ``\textgreater'' indicate the placeholders.}
   \label{fig:task_instructions}
   \vspace{-0.2cm}
\end{figure}

Based on the annotation pipeline introduced in Sec. \textcolor{red}{4} and \cref{sec:annot_pipeline_appendix}, we construct training data for eight tasks in Stage-1 with task-specific instructions and response templates. \cref{fig:task_instructions} shows the examples of task-specific instructions and response templates.

\subsection{Training data organization for Stage-2}
\begin{figure}
  \centering
   \includegraphics[width=0.99\linewidth]{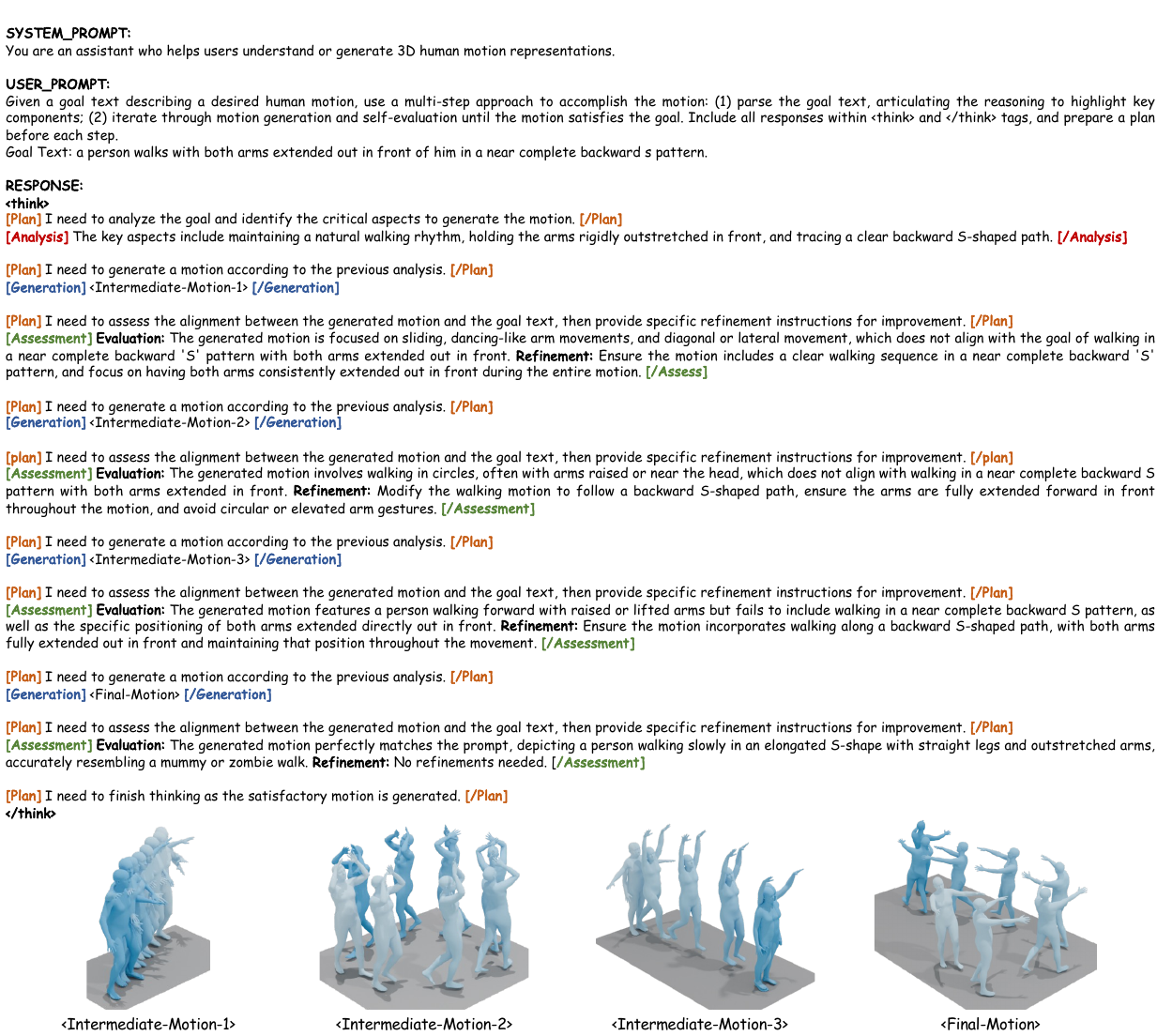}
    \vspace{-0.4cm}
   \caption{\textbf{Visualization of the IRMoGen-CoT data for Stage-2 training.}}
   \label{fig:cot_case}
   \vspace{-0.3cm}
\end{figure}

As discussed in \cref{sec:annot_pipeline_appendix}, each set of negative motion-text pairs includes $N_{neg}$ negative motion-text pairs for each anchor motion-text pair
Based on this, we can build reasoning trajectories that match the IRMoGen-CoT template shown in Fig. \textcolor{red}{3} for the Stage-2 training. Each reasoning trajectories may include at most $N_{neg}$ \emph{intermediate motions} and one \emph{final motion}.
During training, we remove a random number (0 to 4) of the \emph{intermediate motions} together with the corresponding assessment results to ensure the diversity of lengths of reasoning trajectories. We also filter out those intermediate motions with refinement instruction of ``\emph{No refinements needed}'' to ensure only the final motion is perfectly aligned with the goal.
We also provide an example of the Stage-2 training data \cref{fig:cot_case} (including once initial generation and three rounds of refinement).


\subsection{Comparison between the construction of MotionFix \cite{motionfix}} 
While both MotionFix dataset and our data engine use motion encoder to retrieve motions, but different in the outcomes.
MotionFix retrieves most similar motions and provides brief editing prompts. However, our engine retrieves \emph{graded multi-level negatives}, generates \emph{detailed alignment evaluations} and corresponding \emph{refinement instructions} to enable native generation-assessment-refinement loops.

\section{More about Experiment Setups}
\label{sec:more_experiment_settings}
\subsection{More about Dataset}
Following previous works \cite{motiongpt, motiongpt3, motionagent, motionchain, mg-motionllm} on MoLLM, we mainly conduct experiments on HumanML3D dataset \cite{guo}, one of the
largest motion-text pair datasets. It contains 14,616 motion sequences from AMASS \cite{AMASS} and HumanAct12 \cite{guo2020action2motion}, along with 44,970 sequence-level textual descriptions. These descriptions capture the semantic meaning of each motion, making HumanML3D widely used for motion generation and understanding tasks. 
To evaluate the generalizability of our approach, we further compare IRG-MotionLLM with other methods on the KIT-ML \cite{kit-ml} dataset, which contains 3911 motion sequences and 6,278 text descriptions. The motion sequences are selected from the KIT \cite{kit-ml} and CMU \cite{cmu-dataset} datasets.
We follow the official splitting strategy to split the dataset into training, validation and test set.
For Stage3 training, we keep the ratio of text mode paired data to text only data to be approximately 1:1. For fair evaluation, we filter out those prompts that also exist in the testing set (considering them identical after converting to lowercase and removing punctuation).

\subsection{More about Evaluation Metrics}
\textbf{For the Text-to-Motion Generation task}, we follow previous works \cite{guo, motionagent} to report metrics including R-Precision, FID, MM-Dist, and Diversity. Given the ground-truth motion features $f_{gt}$, generated motion features $f_{pred}$, and text features $f_{text}$. Here are details about metrics.

- \textbf{Multimodal Distance} (MM-Dist) is utilized to measure the feature-level distance between the text features and the generated motion feature:  

\vspace{-0.2cm}
\begin{equation}
\text{MM-Dist}=\frac{1}{N}\sum^{N}_{i=1}||f_{pred,i}-f_{text,i}||,
\end{equation}
where $f_{pred,i}$ and $f_{text,i}$ denote the features of the $i$-th generated motion and the corresponding text.

- \textbf{Frechet Inception Distance} (FID) measures the distance of motion features distribution between ground-truth and generated motions, which is calculated FID by
\begin{equation}
\text{FID} = \|\mu_{gt} - \mu_{pred}\|^2 
     - \operatorname{Tr}(
         \Sigma_{gt} + \Sigma_{pred} - 2(\Sigma_{gt} \Sigma_{pred})^{1/2}),
\end{equation}
where $\mu_{gt}$ and $\mu_{pred}$ are the means of $f_{gt}$ and $f_{pred}$. $\Sigma$ is the covariance matrix and Tr denotes the trace of a matrix.

- \textbf{R-Precision} (Top-k) is a text-motion retrieval metric. Given the generated motion and a mini-batch of text descriptions (one ground-truth and others are randomly selected from mismatched descriptions), we rank the distances between the motion and text features to get Top-k accuracy of motion-text. Following the official evaluation protocol \cite{guo}, the mini batch size is set to 32, Euclidean distance is adopted, and we report Top-1, Top-2 and Top-3 results.

- \textbf{Diversity} measures the variance of the whole motion sequences across the dataset. We randomly sample $N_{div}$ pairs of motion and each pair of motion features is denoted by $f_{pred,i}$ and $f^{'}_{pred,i}$. The diversity can be calculated by
\begin{equation}
\text{Diversity}=\frac{1}{N_{div}}\sum^{N_{div}}_{i=1}||f_{pred,i}-f^{'}_{pred,i}||.
\end{equation}
Following official evaluation protocol\cite{guo}, $N_{div}$ is set to 300.

For most of the evaluation, we adopt the official evaluator from HumanML3D \cite{guo}. 
As this evaluator can be influenced by redundant motion representation \cite{mardm}, we also employ a recently proposed evaluator \cite{mardm} for further evaluation. 

\begin{table}[t]
    \scriptsize
    \centering
    \scriptsize
    \caption{\textbf{Hyper-parameters for training IRG-MotionLLM.} Since the KIT-ML dataset (KIT) contains fewer motions than the HumanML3D (H3D) dataset, we correspondingly reduced the training epochs / steps for each stage but keep other hyper-parameters the same.}
    \vspace{-0.3cm}
    \resizebox{0.70\linewidth}{!}{
        \begin{NiceTabular}{l|ccc}
        \toprule
        {IRG-MotionLLM} & Stage-1 & Stage-2 & Stage-3\\
        \midrule
        \multicolumn{4}{l}{\emph{{General Hyper-parameters}}} \\
        Batch Size & 128 & 32 & 32 \\
        \# GPUs (Nvidia L20s) & 32 & 32 & 32\\
        Learning Rate & 1e-5 & 1e-5 & 1e-6\\
        Epochs / Steps (H3D) & 10 epochs & 10 epochs & 900 steps\\
        Epochs / Steps (KIT) & 4 epochs & 4 epochs & 300 steps\\
        \midrule
        \multicolumn{4}{l}{\emph{{GRPO Hyper-parameters}}} \\
        KL weight ($\beta_{kl}$) & - & - & 0.01\\
        Group size ($G$) & - & - & 4 \\
        Max completion length & - & - & 1024 \\
        \bottomrule   
        \end{NiceTabular}
    }
    \label{tab:hyper_param_training}
\end{table}

\vspace{0.1cm}
\textbf{For the Motion-to-Text Caption task}, following \cite{motionagent, motiongpt, motionchain}, we adopt NLP metrics including Bleu\cite{bleu}, Rouge-L\cite{rouge}, CIDEr\cite{cider}, and BertScore\cite{bertscore}. Here are details about metrics.

- {\textbf{BLEU}} \cite{bleu} quantifies the overlap between the generated sentences and the reference sentences by comparing n-grams (contiguous word sequences) shared between them. In this work, we compute the 1-gram- and 4-gram-based BLEU score to evaluate the models.

- {\textbf{CIDEr}} \cite{cider} is originally proposed for image captioning task. It measures the similarity between generated and reference texts using n-gram overlap weighted by TF-IDF. It ensures important but less common phrases are emphasized, rewarding consensus among references while penalizing overly generic outputs.

- {\textbf{ROUGE-L}} \cite{rouge} measures the longest common subsequence (LCS) between the generated caption and the reference captions. Unlike n-gram-based metrics that only consider exact matches of contiguous words, ROUGE-L captures sentence-level structure similarity and allows for non-consecutive word matches as long as they preserve relative order, making it particularly suitable for evaluating fluency and structural consistency in generated text.

- {\textbf{BERTScore}} \cite{bertscore} evaluates natural language generation by comparing the semantic similarity between generated and reference texts using contextual embeddings from pre-trained BERT-like models. Unlike traditional metrics that rely on n-gram overlap, BERTScore computes token-level similarity based on meaning, capturing nuances like synonyms and paraphrasing.

\subsection{More Implementation Details}
\vspace{-0.1cm}
For experiments on the HumanML3D dataset, we build our base model based on the official weights of MotionLLM \cite{motionagent}. For experiments on the KIT-ML dataset, we follow the official implementation (see \cref{sec:more_model_details} and the original paper \cite{motionagent}) of MotionLLM \cite{motionagent} to train the base model for 100 epochs.
The main hyper-parameters for training our IRG-MotionLLM are shown in \cref{tab:hyper_param_training}.
During inference, the Maximum Refinement Constraint $N_{max}$ is set to 4 (see discussions in \cref{sec:inference_IRMoGen}). Note that with proper maximum completion length, after Stage-3 training, only 0.2\% test reasoning trajectories exceed this constraint.
All our experiments are conducted on NVIDIA L20 GPUs.
During inference, TFLOPs increase with the number of refinement rounds: 1.8, 2.4, 3.2, 4.0, and 4.7 for 0 to 4 rounds.

\section{More Experiment Results and Discussions}
\label{sec:more_experiments}

\begin{table}[t]
    \scriptsize
    \centering
    \scriptsize
    \caption{\textbf{Ablation on the data filtering strategy.} The manual checking and filtering step bring  consistent positive impact to the generation performance.}
    \vspace{-0.3cm}
    \resizebox{0.80\linewidth}{!}{
        \begin{NiceTabular}{l|c|ccccc}
        \toprule
        \multirow{2}{*}{{Methods}} & \multirow{2}{*}{w/ filtering} & \multicolumn{3}{c}{R-Precision$\uparrow$} & \multirow{2}{*}{FID$\downarrow$} &  \multirow{2}{*}{MM-Dist$\downarrow$}   \\ 
        \cline{3-5}
        & &Top-1 & Top-2 & Top-3\\
        \midrule
        Ours(S1, w/ IRMoGen) & $\times$ & 0.515 & 0.702 & 0.797 & 0.152 & 2.957\\
        \rowcolor{lightpurple}Ours(S1, w/ IRMoGen) & \checkmark & \textbf{0.522} & \textbf{0.711} & \textbf{0.805} & \textbf{0.135} & \textbf{2.906}\\
        \bottomrule   
        \end{NiceTabular}
    }
    \label{tab:ablation_filtering}
\end{table}

\subsection{Impact of manual check on the LLM-based annotations.}
\label{sec:ablation_manual_check}
As discussed in Sec. \textcolor{red}{4}, we manually filter out the invalid LLM-based annotations prior to model training. Here we conduct a further ablation on our Stage-1 model to evaluate the impact of the filtering strategy. As shown in \cref{tab:ablation_filtering}, the manual checking and filtering step bring consistent positive impacts to the generation performance.

\subsection{More Comparisons and Discussions on Text-to-Motion Generation Task.}
\label{sec:more_comparison_t2m}
In \cref{tab:more_comparison_t2m}, we further compare our IRG-MotionLLM with methods with other frameworks on Text-to-Motion Generation task. 
\textbf{First}, IRG-MotionLLM significantly narrows the gap between MoLLMs \cite{motiongpt,motiongpt2,mg-motionllm,motion-r1} and generation-specific models (\eg, \cite{lamp,mardm,hong2025salad}), achieving comparable results with the advanced mask-modeling-based motion generator (\eg, BAMM \cite{bamm}, MoGenST \cite{mogenst}).  
While there exists {a} gap between our approach and the most recent generation-specific method (\ie, LAMP \cite{lamp} and SALAD \cite{hong2025salad}), we note that it is not a challenge only in MoLLM research \cite{mg-motionllm, motionagent, motion-r1}. 
In the domain of image generation and understanding \cite{aia,jiao2025unitoken,zhang2025are}, current unified VLMs also do not exhibit a clear superiority over separate models in the understanding or generation tasks. 
Even so, we still believe exploring how to deeply link motion understanding with generation beyond simply training model on various independent tasks is highly valuable for both the MoLLM and motion generation domains. 
\textbf{Second}, IRG-MotionLLM achieves comparable performance compared with methods trained on much larger close-source datasets \cite{being_m05,being-m0}. We believe our IRG-MotionLLM can also benefit from larger training data and advanced architectures.

\clearpage
\newpage

\begin{table}[t]
    \scriptsize
    \centering
    \scriptsize
    \caption{\textbf{Compared with existing methods on Text-to-Motion task.} Without specific noted, we use the results reported in the original papers. *: We use the official weights of MotionLLM \cite{motionagent} as our base model for experiments on the HumanML3D dataset. $\S$: We train our base model following the official codebase of MotionLLM \cite{motionagent} on the KIT-ML dataset. 
    $\ddagger$: We scale up the training data in Stage-3 to the similar scale as UniMo \cite{wang2026unimo} and extend the GRPO duration (also see \cref{tab:more_comparison_grpo_model}).
    Models with $\clubsuit$ support both text and motion outputs. We also report 95\% confidence intervals if it is available in the original paper. We gray out results from models trained with much larger close-sourced dataset for fair comparisons.} 
    \vspace{-0.3cm}
    \resizebox{1\linewidth}{!}{
        \begin{NiceTabular}{lc|ccccccc}
        \toprule
        \multirow{2}{*}{{Methods}} &  \multirow{2}{*} {w/ GRPO} & \multicolumn{3}{c}{R-Precision$\uparrow$} & \multirow{2}{*}{FID$\downarrow$} &  \multirow{2}{*}{MM-Dist$\downarrow$} & \multirow{2}{*}{Diversity$\uparrow$}   \\ 
        \cline{3-5}
        &&Top-1 & Top-2 & Top-3\\
        \midrule 
        \multicolumn{8}{c}{\small \emph{\textbf{HumanML3D Dataset}}} \\       

        \multicolumn{9}{l}{\emph{\textbf{Diffusion-based Motion Generator}}} \\
        MDM\cite{mdm}\tiny ICLR'23 & $\times$&$0.320^{\pm.005}$ & $0.498^{\pm.004}$ & $0.611^{\pm.007}$ & $0.544^{\pm.044}$ & $5.566^{\pm.027}$ & $9.559^{\pm.086}$\\
        MLD\cite{mld}\tiny CVPR'23 & $\times$&$0.481^{\pm.003}$ & $0.673^{\pm.003}$ & $0.772^{\pm.002}$ & $0.473^{\pm.013}$ & $3.196^{\pm.010}$ & $\textbf{9.724}^{\pm.082}$\\
        MotionDiffuse\cite{motiondiffuse}\tiny TPAMI'24 &$\times$& $0.491^{\pm.001}$& $0.681^{\pm.001}$ &$0.782^{\pm.001}$ &$0.630^{\pm.001}$ &$3.113^{\pm.001}$ &$9.410^{\pm.049}$\\
        ReMoDiffuse\cite{remodiffuse}\tiny ICCV'23 &$\times$& $0.510^{\pm.005}$ & $0.698^{\pm.006}$ & $0.795^{\pm.004}$ & $0.103^{\pm.004}$
 & \textbf{$2.974^{\pm.016}$} & $9.018^{\pm.075}$\\
        MARDM\cite{mardm}\tiny CVPR'25 & $\times$ & {0.523} & {0.715} & {0.810} & \textbf{0.061} & - & -\\
        SALAD\cite{hong2025salad}\tiny CVPR'25 & $\times$ & $\textbf{0.581}^{\pm.003}$ & $\textbf{0.769}^{\pm.003}$ & $\textbf{0.857}^{\pm.002}$ & $0.076^{\pm.002}$ & $\textbf{2.649}^{\pm.009}$ & $9.696^{\pm.096}$\\
        \midrule
        \multicolumn{9}{l}{\emph{\textbf{Mask-Modeling-based Motion Generator}}} \\
        
        MoMask\cite{momask}\tiny CVPR'24 &  $\times$ & $0.521^{\pm.002}$ & $0.713^{\pm.002}$ & $0.807^{\pm.002}$ & $0.045^{\pm.002}$ & $2.958^{\pm.008}$ & -\\
        MMM\cite{mmm}\tiny CVPR'24 & $\times$& $0.515^{\pm.002}$ & $0.708^{\pm.002}$ &$0.804^{\pm.002}$ &$0.089^{\pm.005}$ &$2.926^{\pm.007}$ & $9.577^{\pm.050}$\\
        BAMM\cite{bamm}\tiny ECCV'24 &  $\times$ & $0.525^{\pm.002}$ &  $0.720^{\pm.003}$ & $0.814^{\pm.003}$ & $0.055^{\pm.002}$ & $2.919^{\pm008}$ & $\textbf{9.717}^{\pm.089}$\\
        MoGenST\cite{mogenst}\tiny NeurIPS'24 &  $\times$  & $0.529^{\pm.003}$ & $0.719^{\pm.002}$ &  $0.812^{\pm.002}$ & ${0.033}^{\pm.001}$  & $2.867^{\pm.006}$ & $9.570^{\pm.077}$ \\
        LAMP\cite{lamp}\tiny ICLR'25 & $\times$ & $\textbf{0.557}^{\pm.003}$ & $\textbf{0.751}^{\pm.002}$ & $\textbf{0.843}^{\pm.001}$ & $\textbf{0.032}^{\pm.002}$ & $\textbf{2.759}^{\pm.007}$ & $9.571^{\pm.069}$\\
        \midrule
        \multicolumn{9}{l}{\emph{\textbf{VQ-based Motion-aware LLM}}} \\       
        \textcolor{gray}{Being-M0}\cite{being-m0}\tiny \textcolor{gray}{ICML'25} \scriptsize &$\times$ &\textcolor{gray}{0.519} & \textcolor{gray}{-} &\textcolor{gray}{0.803} & \textcolor{gray}{0.166} & \textcolor{gray}{2.964} & \textcolor{gray}{-}\\
        \textcolor{gray}{Being-M0.5}$\clubsuit$\cite{being_m05}\tiny \textcolor{gray}{ICCV'25} \scriptsize & $\times$& \textcolor{gray}{0.528} & \textcolor{gray}{-} & \textcolor{gray}{0.815} & \textcolor{gray}{0.141} & \textcolor{gray}{2.953} & \textcolor{gray}{-}\\
        T2M-GPT\cite{t2m-gpt} \tiny {CVPR'23} \scriptsize &  $\times$& $0.492^{\pm.003}$ & $0.679^{\pm.002}$ & $0.775^{\pm.002}$ & $0.141^{\pm.005}$ & $3.121^{\pm.009}$ & $9.722^{\pm.082}$ \\

        MotionGPT$\clubsuit$ \cite{motiongpt}\tiny NeurIPS'23 \scriptsize & $\times$ & $0.492^{\pm.003}$ & $0.681^{\pm.003}$ & $0.733^{\pm.006}$ & $0.232^{\pm.008}$ & $3.096^{\pm.008}$ & $9.528^{\pm.071}$ \\

        MotionChain$\clubsuit$ \cite{motionchain}\tiny ECCV'24 \scriptsize & $\times$ & $0.504^{\pm.003}$ & $0.617^{\pm.002}$ & $0.790^{\pm.003}$ & $0.248^{\pm.009}$ & $3.033^{\pm.010}$ & $9.470^{\pm.075}$ \\
        MG-MotionLLM$\clubsuit$\cite{mg-motionllm}\tiny CVPR'25 \scriptsize & $\times$& $0.516^{\pm.002}$ & $0.706^{\pm.002}$ & $0.802^{\pm.003}$ & $0.303 ^{\pm.010}$ & $2.952^{\pm.009}$ & $9.960^{\pm.073}$ \\
        ViMoRAG\cite{vimorag}  \tiny {NeurIPS'25} \scriptsize &  $\times$& $0.452^{\pm.014}$ & $0.655^{\pm.014}$ & $0.764^{\pm.013}$ & $0.131^{\pm.073}$ & $3.146^{\pm.015}$ & $9.424^{\pm.001}$\\
        MotionGPT-2(8B)$\clubsuit$ \cite{motiongpt2}\tiny ArXiv'24 & $\times$ & $0.496^{\pm.002}$ & $0.691^{\pm.003}$ & $0.782^{\pm.004}$ & $0.191^{\pm.004}$ & $3.080^{\pm.013}$ & $9.860^{\pm.026}$ \\
        MotionGPT-2(2B)$\clubsuit$\cite{motiongpt2}\tiny ArXiv'24 &  $\times$ & 0.436 & 0.600 & 0.697 & 0.228 & 3.589 & \textbf{10.081} \\
        \scriptsize MotionLLM \cite{motionagent}\tiny {ICLR'25} \scriptsize &  $\times$ & $0.515^{\pm.004}$ & $0.691^{\pm.003}$ & $0.801^{\pm.004}$ & $0.230^{\pm.009}$ & $2.967^{\pm.020}$ & $9.908^{\pm.102}$ \\
        \scriptsize MotionLLM* \cite{motionagent} \tiny {ICLR'25}  &  $\times$& ${0.496}^{\pm.002}$ & ${0.684}^{\pm.002}$ & ${0.782}^{\pm.002}$ & ${0.198}^{\pm.006}$ & ${3.029}^{\pm.007}$ & ${9.864}^{\pm.007}$\\
        \rowcolor{lightpurple}\scriptsize Ours (S1, w/ IRMoGen)$\clubsuit$   &  $\times$&$0.522^{\pm.003}$& $0.711^{\pm.002}$& $0.805^{\pm.002}$ & $0.135^{\pm.005}$	& $2.906^{\pm.009}$ & $9.740^{\pm.080}$\\
        \rowcolor{lightpurple}\scriptsize Ours (S2, w/ IRMoGen)$\clubsuit$ &  $\times$& ${0.526}^{\pm.002}$ & ${0.717}^{\pm.002}$ & ${0.810}^{\pm.002}$ & $\textbf{0.111}^{\pm.005}$ & ${2.885}^{\pm.008}$& $9.819^{\pm.088}$\\
        Motion-R1$\clubsuit$ \cite{motion-r1}\tiny ICLR'26 \scriptsize &  $\checkmark$ & $0.515^{\pm.003}$ & $0.719^{\pm.002}$ & $0.818^{\pm.002}$ & ${0.201}^{\pm.004}$ & $2.854^{\pm.010}$ & ${10.026}^{\pm.075}$ \\
        UniMo$\clubsuit$ \cite{wang2026unimo}\tiny AAAI'26 \scriptsize &  $\checkmark$ & ${0.539}^{\pm.003}$ & ${0.738}^{\pm.002}$ & ${0.831}^{\pm.002}$ & ${0.177}^{\pm.004}$ & ${2.768}^{\pm.010}$ &  ${10.042}^{\pm.076}$\\
        \rowcolor{lightpurple}\scriptsize Ours (S3, w/ IRMoGen)$\clubsuit$ &  $\checkmark$ & ${0.535}^{\pm.002}$ & ${0.725}^{\pm.002}$ & ${0.820}^{\pm.002}$ & ${0.242}^{\pm.006}$ & ${2.785}^{\pm.006}$ &  ${9.900}^{\pm.094}$\\
        \rowcolor{lightpurple}\scriptsize Ours$\ddagger$ (S3, w/ IRMoGen)$\clubsuit$ &  $\checkmark$ & $\textbf{0.564}^{\pm.002}$ & $\textbf{0.754}^{\pm.002}$ & $\textbf{0.841}^{\pm.002}$ & ${0.208}^{\pm.003}$ & $\textbf{2.628}^{\pm.005}$ &  ${9.883}^{\pm.097}$\\
        \midrule
        \multicolumn{8}{c}{\emph{\small \textbf{KIT-ML Dataset}}} \\       
        \multicolumn{9}{l}{\emph{\textbf{Diffusion-based Motion Generator}}} \\
        MDM\cite{mdm}\tiny ICLR'23 & $\times$& - & - & $0.396^{\pm.004}$ & $0.497^{\pm.021}$ & $9.191^{\pm.022}$ & $10.85^{\pm.109}$\\
        MLD\cite{mld}\tiny CVPR'23 & $\times$  & $0.390^{\pm.008}$ & $0.609^{\pm.008}$ & $0.734^{\pm.007}$ & $0.404^{\pm.027}$ & $3.204^{\pm.027}$ & $10.800^{\pm.117}$\\
        MotionDiffuse\cite{motiondiffuse}\tiny TPAMI'24 &$\times$& $0.417^{\pm.004}$ & $0.621^{\pm.004}$ & $0.739^{\pm.004}$ & $1.954^{\pm.062}$ & $2.958^{\pm.005}$ & $\textbf{11.100}^{\pm.143}$\\
        ReMoDiffuse\cite{remodiffuse}\tiny ICCV'23 &$\times$& ${0.427}^{\pm.014}$ & ${0.641}^{\pm.004}$ & ${0.765}^{\pm.055}$ & $\textbf{0.155}^{\pm.006}$ & ${2.814}^{\pm.012}$ & $10.800^{\pm.105}$\\
        SALAD\cite{hong2025salad}\tiny CVPR'25 & $\times$ & $\textbf{0.477}^{\pm.006}$ & $\textbf{0.711}^{\pm.005}$ & $\textbf{0.828}^{\pm.005}$ & $0.296^{\pm.012}$ & $\textbf{2.585}^{\pm.016}$ & $11.097^{\pm.095}$\\
        \midrule
        \multicolumn{8}{l}{\emph{\textbf{Mask-Modeling-based Motion Generator}}} \\
        
        MoMask\cite{momask}\tiny CVPR'24 &  $\times$ & $0.433^{\pm.007}$ & $0.656^{\pm.005}$ & $0.781^{\pm.005}$ & $0.204^{\pm.011}$ & $2.779^{\pm.022}$ & -\\
        MMM\cite{mmm}\tiny CVPR'24 & $\times$&  $0.404^{\pm.005}$ & $0.621^{\pm.005}$ & $0.744^{\pm.004}$ & $0.316^{\pm.028}$ & $2.977^{\pm.019}$ & $10.910^{\pm.101}$\\
        BAMM\cite{bamm}\tiny ECCV'24 &  $\times$ & $0.438^{\pm.009}$ & $0.661^{\pm.009}$ & $0.788^{\pm.005}$ & $0.183^{\pm.013}$ & $2.723^{\pm.026}$ & $\textbf{11.008}^{\pm.094}$\\
        MoGenST\cite{mogenst}\tiny NeurIPS'24 &  $\times$ & $0.445^{\pm.006}$ & $0.671^{\pm.006}$ &  $0.797^{\pm.005}$ & $0.143^{\pm.004}$ & $2.711^{\pm.024}$ & $10.918^{\pm.090}$ \\
        LAMP\cite{lamp}\tiny ICLR'25 & $\times$ & $\textbf{0.479}^{\pm.006}$ & $\textbf{0.691}^{\pm.005}$ & $\textbf{0.826}^{\pm.005}$ & $\textbf{0.141}^{\pm.013}$ & $\textbf{2.704}^{\pm.018}$ & ${10.929}^{\pm.101}$ \\
        \midrule
        \multicolumn{9}{l}{\emph{\textbf{VQ-based Motion-aware LLM}}} \\       
        T2M-GPT\cite{t2m-gpt} \tiny {CVPR'23} \scriptsize & $\times$& $0.416^{\pm.006}$ & $0.627^{\pm.006}$ & $0.745^{\pm.006}$ & $0.514^{\pm.029}$ & $3.007^{\pm.023}$ & $10.921^{\pm.108}$ \\
        MotionGPT$\clubsuit$ \cite{motiongpt}\tiny NeurIPS'23 \scriptsize & $\times$ & $0.366^{\pm.005}$ & $0.558^{\pm.004}$ & $0.680^{\pm.005}$ & $0.510^{\pm.016}$ & $3.527^{\pm.021}$ & $10.350^{\pm.084}$\\
        MotionGPT-2(2B)$\clubsuit$ \cite{motiongpt2}\tiny ArXiv'24 \scriptsize & $\times$ & $0.364$ & $0.581$ & $0.699$ & $1.063$ & $3.424$ & $10.603$\\
        \scriptsize MotionLLM$\S$ \cite{motionagent} \tiny {ICLR'25} &  $\times$& $0.416^{\pm.004}$ & $0.637^{\pm.004}$ & $0.746^{\pm.003}$ & $0.560^{\pm.025}$ & $3.001^{\pm.011}$ & $\textbf{11.209}^{\pm.135}$\\
        \rowcolor{lightpurple}\scriptsize Ours (S1, w/ IRMoGen)$\clubsuit$  &   $\times$& ${0.425^{\pm.005}}$ & $0.640^{\pm.004}$ & $0.754^{\pm.002}$ & $0.497^{\pm.021}$ & ${2.877}^{\pm.010}$ & $11.114^{\pm.095}$\\
        \rowcolor{lightpurple}\scriptsize Ours (S2, w/ IRMoGen)$\clubsuit$ &   $\times$&  $0.419^{\pm.006}$ & ${0.642}^{\pm.006}$ & ${0.763}^{\pm.005}$ & ${0.389}^{\pm.014}$ & $2.908^{\pm.026}$ & $11.096^{\pm.121}$ \\
        Motion-R1$\clubsuit$ \cite{motion-r1}\tiny ICLR'26 \scriptsize &   $\checkmark$ & $0.431^{\pm.003}$ & $0.638^{\pm.002}$ & $0.761^{\pm.003}$ & $\textbf{0.287}^{\pm.004}$ & $3.196^{\pm.040}$ & $10.875^{\pm.052}$ \\
        \rowcolor{lightpurple}\scriptsize Ours (S3, w/ IRMoGen)$\clubsuit$ &   $\checkmark$ & $\textbf{0.445}^{\pm.005}$ & $\textbf{0.681}^{\pm.003}$ & $\textbf{0.781}^{\pm.004}$ & ${0.432}^{\pm.013}$ & $\textbf{2.740}^{\pm{.017}}$ & ${11.115}^{\pm .086}$\\
        \bottomrule   
        \end{NiceTabular}
    }
    
    \label{tab:more_comparison_t2m}
\end{table}

\clearpage
\newpage

\begin{table}[t]
    \scriptsize
    \centering
    \scriptsize
    \caption{\textbf{Comparisons with agentic (multi-agent) pipelines}. *: We use the official weights of MotionLLM \cite{motionagent} as the base model. Similar to our Stage-1 model, Asse.Expert and Ref.Expert are independently trained from the base model to specifically perform Motion Refinement Instructing and Instruction-guided Motion Refinement tasks, respectively. See more details in \cref{sec:unified_vs_agentic}.}
    \vspace{-0.3cm}
    \resizebox{1\linewidth}{!}{
        \begin{NiceTabular}{l|l|c|c|cccccc}
        \toprule
        &\multirow{2}{*}{{Methods}} & \multirow{2}{*}{\# Param} & Multi- & \multicolumn{3}{c}{R-Precision$\uparrow$} & \multirow{2}{*}{FID$\downarrow$} &  \multirow{2}{*}{MM-Dist$\downarrow$} & \multirow{2}{*}{Diversity$\uparrow$}   \\  
        \cline{5-7}
        &&&Agent?&Top-1 & Top-2 & Top-3\\
        \midrule
        1&MotionLLM*\cite{motionagent}+Asse.Expert+Ref.Expert & 6B & $\checkmark$ & 0.508 & 0.700 & 0.796 & 0.194 & 2.958 & 9.888\\
        2&MotionLLM*\cite{motionagent}+Ours(S1) & 4B & $\checkmark$& 0.517 & 0.709 & 0.806 & 0.161 & 2.906 & 9.811 \\
        \midrule
        3&\scriptsize MotionLLM* \cite{motionagent} & 2B & $\times$ & ${0.496}$ & ${0.684}$ & ${0.782}$ & ${0.198}$ & ${3.029}$ & ${9.864}$\\
        4&\scriptsize Ours (S1, w/o IRMoGen) & 2B &  $\times$&$0.504$& $0.693$& $0.785$ & $0.141$	& $3.018$ & $9.814$\\
        5&\scriptsize Ours (S1, w/ IRMoGen) & 2B &  $\times$&$0.522$& $0.711$& $0.805$ & $0.135$	& $2.906$ & $9.740$\\
        \bottomrule   
        \end{NiceTabular}
    }
    \label{tab:comparison_agentic}
\end{table}


\subsection{Unified Architecture v.s. Multi-Agent Pipeline}
\label{sec:unified_vs_agentic}
In Sec. \textcolor{red}{5.2}, we conduct ablation studies to show that training a model jointly with motion assessment and refinement tasks enhances the text-motion alignment, and interweaving motion generation with assessment and refinement improves the final generation results. However, one may raise the questions that \emph{whether training a unified model is necessary, or is it possible to observe similar improvement with a multi-agent pipeline} (\ie, improving the generated motion from a base model with independent motion assessment and refinement experts)?

To address this, we conduct a further evaluation to compare the unified architecture with multi-agent pipelines. Specifically, we use MotionLLM \cite{motionagent} as the base model, and consider the following two multi-agent pipelines: 
(1) \emph{MotionLLM+Asse.Expert+Ref.Expert}: A motion assessment expert (Asse.Expert) and a motion refinement expert (Ref.Expert) are trained upon the base model to specifically perform Motion Refinement Instructing and Instruction-guided Motion Refinement, respectively; 
(2) \emph{MotionLLM+Ours(S1)}: We directly using our Stage-1 model to perform Motion Refinement Instructing and Instruction-guided Motion Refinement. During inference, the agentic pipelines follow the similar paradigm shown in \cref{fig:stage1_IRMoGen}, but use various models to perform different tasks. 

From results in \cref{tab:comparison_agentic}, we have the following observations. 
\textbf{First}, comparing Row-3,4 with Row-1,2,5, introducing motion assessment and refinement brings consistent improvement regardless of the type of assessor and refiner.
\textbf{Second}, comparing Row-1 and Row-2, using our Stage-1 model achieves better performance than independent assessment and refinement experts, suggesting that jointly learning assessment, refinement enhances text-motion alignment, resulting in better motion-centric reasoning ability. 
\textbf{Third}, comparing Row-2 and Row-5, the agentic pipeline achieves close performance with the unified model as they use the same assessor and refiner. However, the unified strategy are more friendly to computational cost with less parameters (2B) compared to the multi-agent pipeline (4B).

\begin{figure}[t]
  \centering
   \includegraphics[width=0.50\linewidth]{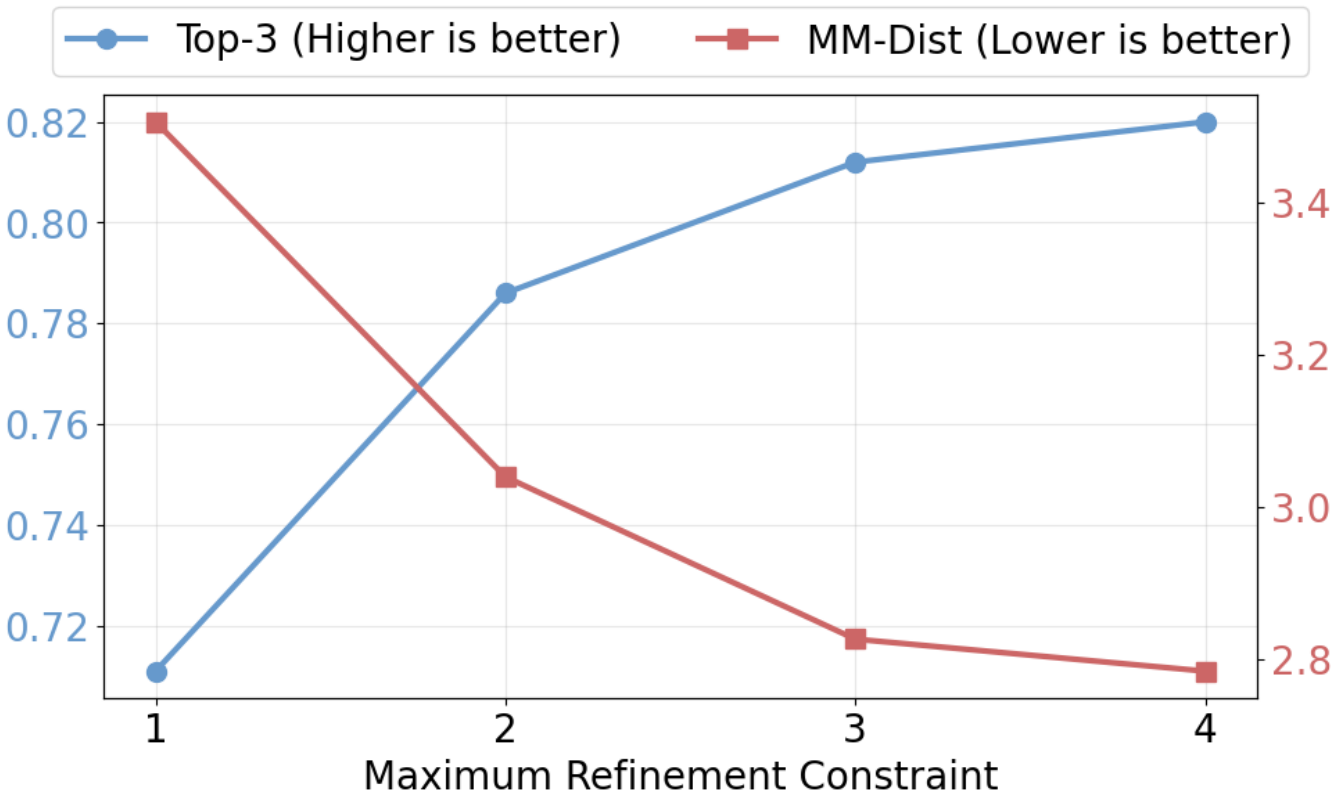}
    \vspace{-0.3cm}
   \caption{\textbf{Ablation studies on the maximum rounds of refinement on the HumanML3D dataset.} Allowing model to perform more rounds of refinement brings improvement on the alignment between generated results and the goal text.}
   \label{fig:ablation_multi_refinement}
   \vspace{-0.3cm}
\end{figure}

\subsection{Impact of multiple round of refinement.}
\label{sec:ablation_multi_round_refinement}
We dive deeper into the multi-round refinement process to show how the round of refinement {affects} the generation performance. As shown in \cref{fig:ablation_multi_refinement}, we constrain the maximum refinement rounds (see discussion in \cref{sec:inference_IRMoGen}) of our Stage-3 model and visualize the performance on text-motion alignment metrics. By performing more rounds of refinement, the alignment between the generated motion and the goal text is consistently improved.

\subsection{Robust IRMoGen or Trajectory Dependency?} 
\label{sec:discussion_rebustness}
Previous studies \cite{roh2025breakthechain,gan2024reasoning,lam2025codecrash} reveal that reasoning LLMs show vulnerability under adversarial attacks, \ie, the reasoning LLMs suffer from significantly performance drop when introducing small perturbations to the prompts or previous CoT. These studies motivate us to raise a question: {\emph{Does IRG-MotionLLm perform robust IRMoGen or only response along a fixed reasoning trajectory?}}

\begin{figure}[t]
  \centering
   \includegraphics[width=0.99\linewidth]{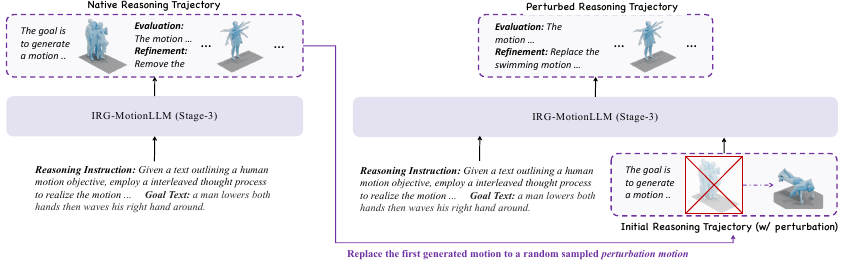}
   \caption{\textbf{Inference pipeline for robustness evaluation.} Given the reasoning instruction and a goal text, we first obtain the native reasoning trajectory of our IRG-MotionLLM. After that, we replace the first generated motion in the native reasoning trajectory to a random sampled perturbation motion, then feed the perturbed initial trajectory (containing only the goal analysis and perturbation motion) together with the instruction and goal back to the model for further reasoning.}
   \label{fig:robustness_pipeline}
   \vspace{-0.3cm}
\end{figure}

\begin{table}[t]
    \scriptsize
    \centering
    \scriptsize
    \caption{\textbf{Robustness evaluation of Interleaved Reasoning on the HumanML3D dataset.} Introducing the \emph{perturbation motions} into the reasoning trajectory (w/ perturb.) only brings slight performance drop, which showcases the robustness of IRMoGen performed by our method. ``perturb. only'': use the \emph{perturbation motions} for evaluation.}
    \vspace{-0.3cm}
    \resizebox{0.8\linewidth}{!}{
        \begin{NiceTabular}{l|c|ccccc}
        \toprule
        \multirow{2}{*}{{Methods}} &  \multirow{2}{*}{Perturb. Position} & \multicolumn{2}{c}{R-Precision$\uparrow$} & \multirow{2}{*}{FID$\downarrow$} &  \multirow{2}{*}{MM-Dist$\downarrow$} & \multirow{2}{*}{Diversity$\uparrow$}   \\  
        \cline{3-4}
        &&Top-1 & Top-3\\
        \midrule
        perturb. only & - &0.029 & 0.084 & - & 9.957 & 9.815 \\
        \midrule
        \scriptsize Ours w/ perturb.  & Random & 0.527 & 0.807 & 0.246 & 2.850  & 9.781\\
        \scriptsize Ours w/ perturb.  & First & 0.526 & 0.810 & {0.227} & 2.826 & 9.874\\
         \scriptsize Ours  & - & {0.535} & {0.820} & 0.242 & {2.785} & {9.900}\\
        \bottomrule   
        \end{NiceTabular}
    }
    
    \label{tab:ablation_perturb}
\end{table}

To answer this question, as shown in \cref{fig:robustness_pipeline} we introduce perturbations to the native interleaved reasoning trajectory by replacing \textbf{the first generated motion} with a random sampled one (noted as \textbf{\emph{perturbation motion}}) from the dataset. After that, we feed the \emph{reasoning instruction}, \emph{goal text}, together with the initial reasoning trajectory (containing only the \emph{goal analysis} and \emph{perturbation motion}) back our IRG-MotionLLM to perform further reasoning. 
We evaluate how the \emph{perturbation motion} could influence the generation performance of our Stage-3 model.

Moreover, in the training data for Stage-2, we assume that the intermediate motions show progressive improvement. Since a more common and realistic scenario is that the reasoning process may occasionally deviate from the ground truth and then correct itself later, such assumption may bring another question: \emph{Is our model robust for the occasional deviations?} To answer this question, we extend the robustness evaluation mentioned above by further replacing the generated motion with a perturbation at a \textbf{random position}.

As shown in \cref{tab:ablation_perturb}, introducing \emph{perturbation motions} to the reasoning trajectories only brings slight performance drop to our IRG-MotionLLM, which showcases the robustness of IRMoGen performed by our method. We provide qualitative visualizations on this study in \cref{sec:more_visualizations}.

\subsection{Stage-3 Training: Unleashing the capacity or Hacking the evaluator?}
\label{sec:discussion_grpo_hacking}
In our Stage-3 training stage, we conduct GRPO-based RL tuning to further enhance the generation performance of our method. We follow previous works \cite{instructmotion-arxiv,motion-r1} to select the reward model and strictly ensure there are no identical prompts between the RL data and the test set. However, as we use the official evaluator \cite{guo} as the reward model, an extreme question may be raised: \emph{Does GRPO-based RL tuning in Stage-3 unleash the potential capacity or simply hack the evaluator?}

To answer this question, as shown in Tab. \textcolor{red}{5} of the main paper, we further compare our IRG-MotionLLM with existing methods by using a newly proposed T2M evaluator \cite{mardm}. By performing Stage-3 training, our method achieves the improvement consistent with the results with original evaluator \cite{guo}, and achieves SoTA performance compared with existing methods with various frameworks. These results suggest that Stage-3 training goes far beyond hacking the evaluator and unleashes the potential capacity of our IRG-MotionLLM.

\subsection{Functioning IRG-MotionLLM as a Text-Motion Reward Model.}
\label{sec:irg_motionllm_as_reward}
So far, we have shown that IRG-Motion performs robust evaluation on the alignment between the generated motion and goal text. However, a question remain unresolved: \emph{How can the assessment ability of IRG-MotionLLM benefit existing motion generator?} The intuition behind this question is that if our IRG-MotionLLM is capable to perform \emph{high-quality} and \emph{transferable} motion assessment, the assessment signal can be used to enhance another motion generator.

To evaluate IRG-MotionLLM's motion assessment capacity, inspired by existing works on Reinforcement Learning from AI Feedback (RLAIF) \cite{atom,pappa2024modipo,VQAScore,zhang2024generative}, we function our IRG-MotionLLM as a new reward model to evaluate the alignment between generated motion and the goal text, and use it to further enhance the existing motion generator. We then compare IRG-MotionLLM with other possible reward models \cite{achiam2023gpt,motionpatch} on AToM-general Benchmark \cite{atom}.

\begin{table}[t]
    \scriptsize
    \centering
    \caption{\textbf{Evaluations with other possible reward models on AToM-general benchmark \cite{atom}.} Following \cite{atom}, we use MotionGPT \cite{motiongpt} (initialized with the official weights) as the base model, and modify AToM framework with different possible reward models.}
    \vspace{-0.3cm}
    \resizebox{0.8\linewidth}{!}{
        \begin{NiceTabular}{l|c|ccccc}
        \toprule
        \multirow{2}{*}{{Methods}} & Reward& \multicolumn{2}{c}{R-Precision$\uparrow$} & \multirow{2}{*}{FID$\downarrow$} &  \multirow{2}{*}{MM-Dist$\downarrow$} & \multirow{2}{*}{Diversity$\uparrow$}   \\  
        \cline{3-4}
        & Models & Top-1  & Top-3\\
        \midrule
        MotionGPT & - & 0.407 & 0.661 & 0.188 & 3.998 & \textbf{9.452}\\
        \midrule
        MotionGPT & GPT4Vision\cite{achiam2023gpt} & 0.410 & 0.669 & 0.177 & 3.943 & 9.401\\
        + RLAIF & MotionPatch\cite{motionpatch} & 0.414	& 0.683 & 0.164 & 3.792 & 9.364\\
         & \rowcolor{lightpurple} Ours & \textbf{0.429} & \textbf{0.698} & \textbf{0.127} & \textbf{3.700} & 9.401\\
        \bottomrule   
        \end{NiceTabular}
    }
    
    \label{tab:comparison_atom}
\end{table}

\vspace{0.1cm}
\noindent\textbf{Text-Motion Alignment Score Calculation}. Inspired by VQAScore \cite{VQAScore}, we calculate the text-motion alignment score based on the output probability of tokens that contain evaluation signals. 
Specifically, we first ask the model to perform native interleaved reasoning. After that, we use the same strategy as introduced in \cref{sec:discussion_rebustness} and \cref{fig:robustness_pipeline} to insert the generated motion into the reasoning trajectory. 
After that, we feed the reasoning instruction, goal text goal analysis, generated motion together with step plans and assessment prefix (``\emph{Evaluation: The generated motion}'') to the model. Finally, we compute the output probability of ``\emph{perfectly}'' on the first output token, and use it as the alignment score. The reason behind this strategy is that in Stage-2, model is trained to respond with ``\emph{Evaluation: The generated motion perfectly aligns with ...}'' only when the given motion and goal text are perfectly aligned. 

\vspace{0.1cm}
\noindent\textbf{Building Preference Pairs \& Optimization}. Following AToM \cite{atom}, we use MotionGPT \cite{motiongpt} as the base model. We use the same goal text prompts as provided in official repository of AToM \cite{atom} to generate motions and build 3600 preference pairs based on the text-motion alignment scores. The optimization details are kept the same as the official implementation of AToM \cite{atom}.

\vspace{0.1cm}
\noindent\textbf{Comparisons}. 
We compare our IRG-MotionLLM with GPT-4o \cite{gpt-4o} (the reward model used in AToM \cite{atom}) and MotionPatch \cite{motionpatch} (a SoTA text-motion retrieval model). As shown in \cref{tab:comparison_atom}, functioning IRG-MotionLLM as a text-motion reward model clearly enhances the baseline model performance on text-to-motion generation. Additionally, IRG-MotionLLM outperforms other possible reward models, highlighting the \textbf{quality} and \textbf{transferability} of its motion assessment ability.

\begin{table}[t]
    \centering
    \scriptsize
    \caption{\textbf{Comparison on text-based motion editing benchmark}\cite{motionfix}. We adapt our model for motion editing by finetuning it on MotionFix dataset(+FT) and report Generated-Target Retrieval results. TMED$\dagger$ is proposed together with MotionFix \cite{motionfix} and reimplemented in the same skeleton format as ours. We source the results from \cite{motionlab}}
    \vspace{-0.3cm}
    \resizebox{0.6\linewidth}{!}{
        \begin{NiceTabular}{l|cccc}
        \toprule
        Methods & R\text{@}1$\uparrow$ & R\text{@}2$\uparrow$ & R\text{@}3$\uparrow$ & AvgR$\downarrow$\\ 
        \midrule
        TMED$\dagger$ \cite{motionfix} & 38.7 & 50.6 &  62.2 & 4.2\\
        \midrule
        MotionLLM*\cite{motionagent}+FT & 39.3 & 54.2 & 63.8 & 5.0\\
        \rowcolor{lightpurple} Ours+FT & \textbf{45.3} & \textbf{61.7} &\textbf{70.5} & \textbf{4.0}\\
        \bottomrule   
        \end{NiceTabular}
    }
    \label{tab:comparison_motionfix}
\end{table}

\subsection{Adapting IRG-MotionLLM to Motion Editing Task.}
\label{sec:irg_motionllm_to_editing}
A remaining question from the other side is that \emph{whether learning assessment and refinement could benefit motion editing task} (\ie, a similar task to motion refinement). To answer this question, we conduct further evaluation on MotionFix \cite{motionfix}, a text-based motion editing benchmark. We adapt our model for motion editing by finetuning it on MotionFix training data (+FT) and compare it with the baseline models \cite{motionfix, motionagent}. The generated-to-target retrieval results in \cref{tab:comparison_motionfix} shows the strong performance of IRG-MotionLLM among the baselines, further revealing the cross-task synergy bought by learning IRMoGen.

\subsection{User Studies}
\label{sec:user_study}
To further evaluate the effectiveness of our method, we employ 4 users to finish the following user studies:

\noindent \textbf{Semantic Alignment \& Naturalness}. We randomly sample 80 prompts and ask users to score (scale:1-4) the \emph{semantic} alignment and \emph{naturalness} of final motions generated by different methods. The average scores are shown in \cref{tab:user_study_overall}. Stage-3 training does not degrade the naturalness of the final motions (comparable with Stage-2 model and base model), but significantly improved semantic alignment with the goal texts

\noindent \textbf{Effectiveness of Multi-Round Refinement}.  To assess multi-round refinement by our model, we first randomly sample 30 pairs of motions before and after each refinement round (120 pairs in total) and ask users to determine whether a refinement is effective (without motion degradation). The average effective rates across refinement rounds 1-4 are: 87\%, 80\%, 77\% and 77\%. 

We further ask the users to compare 50 randomly sampled initial-final motions pairs. The final motions win in 60\%, tie in 28\%, and loss in 12\% of cases.  
While the multi-round refinement is not perfect, it offers a high probability of generating high-quality results (only 12\% degradation).

\begin{table}[t]
    \centering
    \scriptsize
    \caption{\textbf{User study results on Semantic Alignment \& Naturalness}. *: We use the official weights of MotionLLM \cite{motionagent}.}
    \vspace{-0.3cm}
    \resizebox{0.5\linewidth}{!}{
        \begin{NiceTabular}{l|cc}
        \toprule
        Methods & Semantic$\uparrow$ & Naturalness$\uparrow$\\ 
        \midrule
        MotionLLM*\cite{motionagent} & 2.6 & 3.2\\
        \rowcolor{lightpurple} Ours (S2) & 2.8 & 3.4\\
        \rowcolor{lightpurple} Ours (S3) &  \textbf{3.4} & \textbf{3.6}\\
        \bottomrule   
        \end{NiceTabular}
    }
    \label{tab:user_study_overall}
\end{table}

\subsection{More Qualitative Results}
\label{sec:more_visualizations}
\begin{figure}[t]
  \centering
   \includegraphics[width=0.99\linewidth]{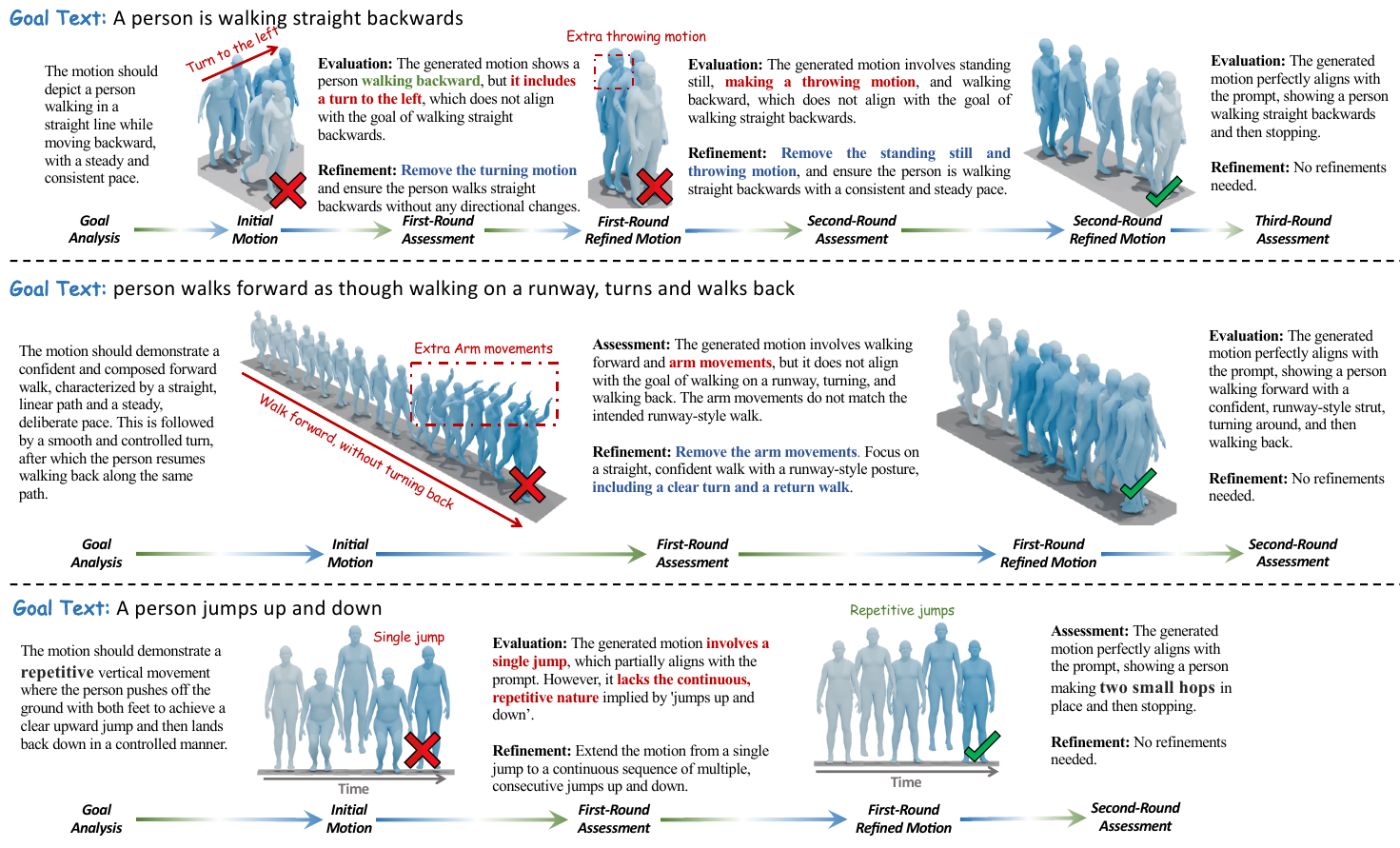}
    \vspace{-0.3cm}
   \caption{\textbf{More native interleaved reasoning trajectories of IRG-MotionLLM.} Our method performs multi-round interleaved reasoning to capture the mis-alignment of the previous generated motion and finally generate a satisfactory motion. The planning steps are removed. {Zoom in for best view.}}
   \label{fig:vis_appendix_1}
   \vspace{-0.2cm}
\end{figure}

\noindent\textbf{More native reasoning trajectories}. \cref{fig:vis_appendix_1} shows the native reasoning trajectories (the planning steps are removed) of IRMoGen performed by our IRG-MotionLLM, showcasing its capacities of performing multi-round interleaved reasoning to capture the mis-alignment of the previous generated motion and finally generating a satisfactory motion. 

\begin{figure}[t]
  \centering
   \includegraphics[width=0.99\linewidth]{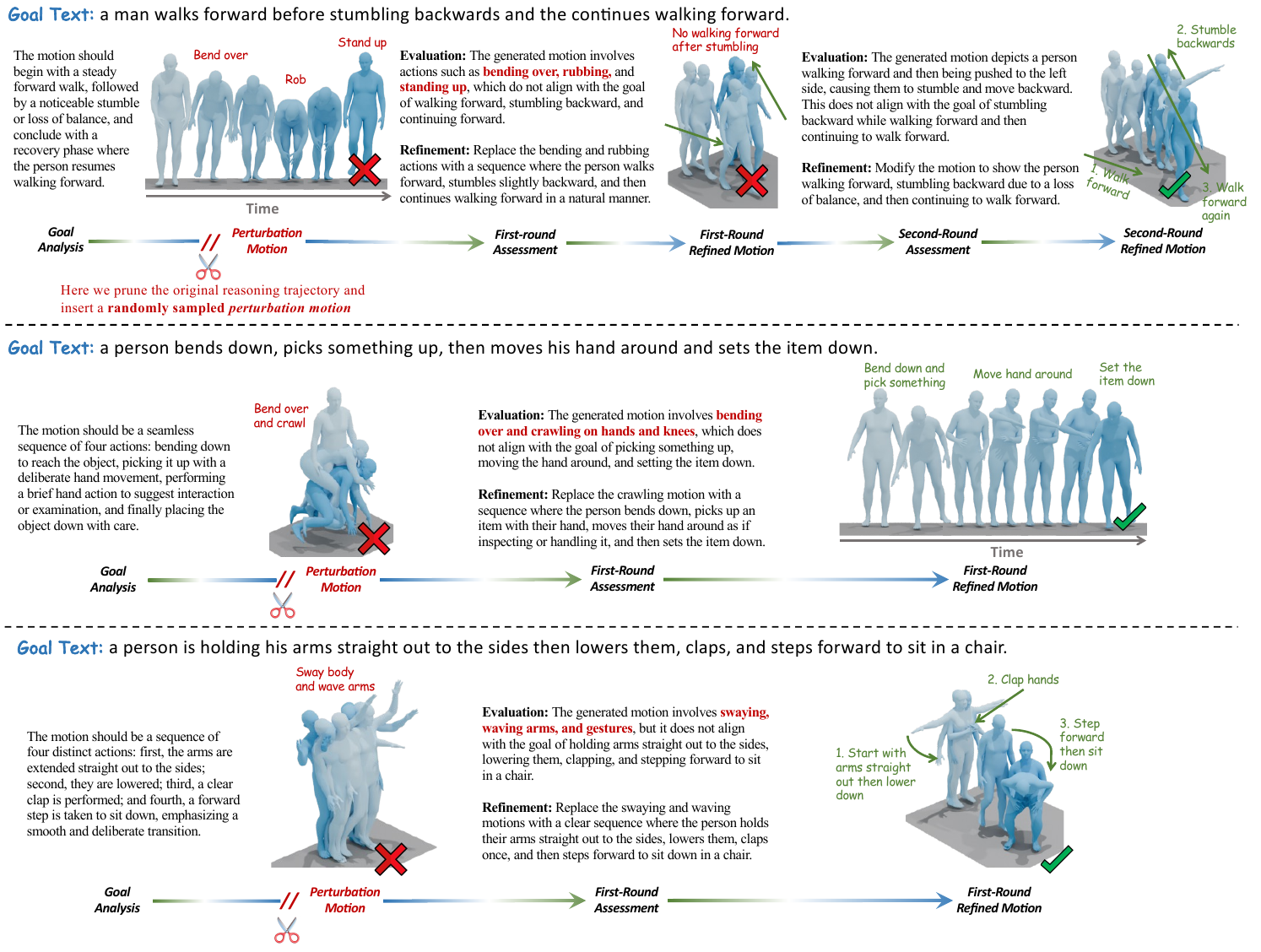}
    \vspace{-0.4cm}
   \caption{\textbf{Qualitative results on robustness evaluation.} Our IRG-MotionLLM can also perform accurate motion assessment on randomly sampled \emph{perturbation motions}, and refine the motion until it has reached the goal.}
   \label{fig:vis_appendix_robustness}
   \vspace{-0.3cm}
\end{figure}

\vspace{0.2cm}
\noindent\textbf{More Robustness Analysis}. 
As discussed in \cref{sec:discussion_rebustness}, we insert \emph{perturbation motions} into the native reasoning trajectory to evaluate the robustness of our IRG-MotionLLM. Here we provide more qualitative results on this study. As shown in \cref{fig:vis_appendix_robustness}, even break the native reasoning trajectory, our IRG-MotionLLM can still perform accurate assessment on the perturbation motion, and refine the motion until it has reached the goal.

\begin{figure}[t]
  \centering
   \includegraphics[width=0.99\linewidth]{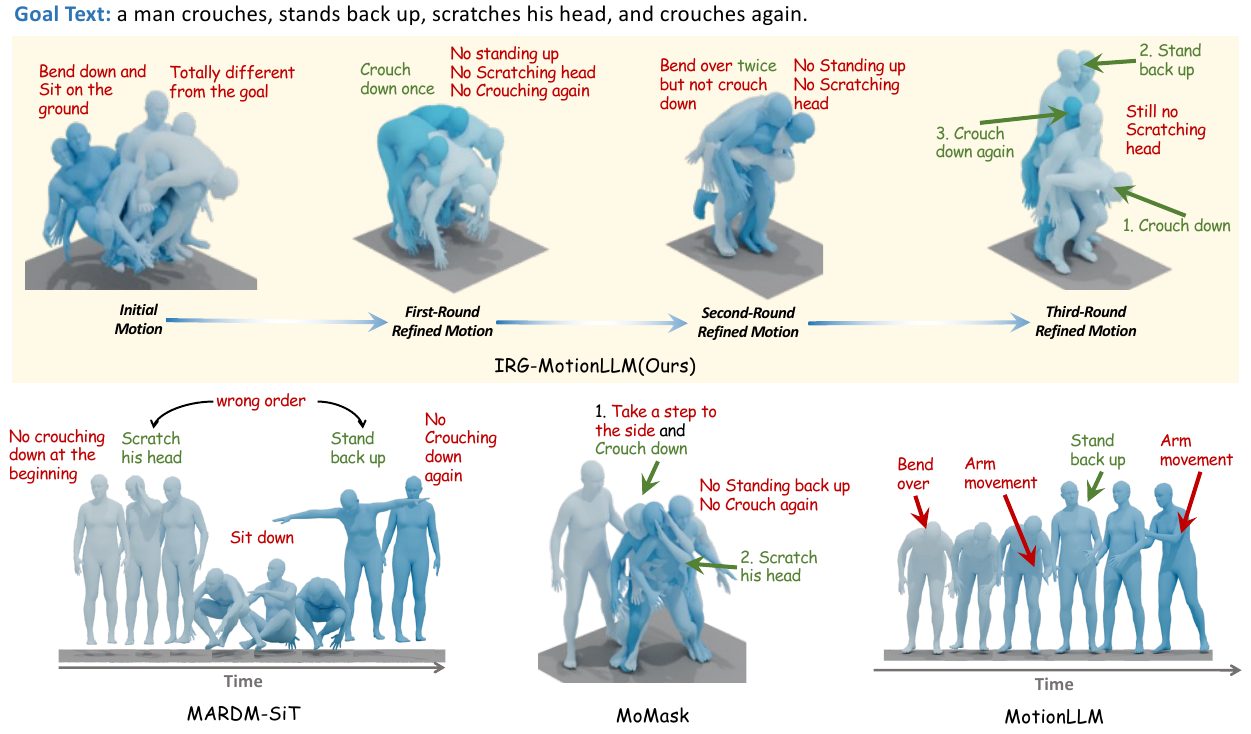}
    \vspace{-0.4cm}
   \caption{\textbf{Our IRG-MotionLLM may fail on highly complex goal text.} Although our IRG-MotionLLM produces a three-round refined motion which is much more aligned with the goal than previous generated motions, it still misses the action of ``\emph{scratching his head}''. However, we note that generating perfect motion to such complex goal text is also challenging for existing methods \cite{momask, mardm, motionagent}.}
   \label{fig:vis_appendix_failure}
\end{figure}

\vspace{0.2cm}
\noindent\textbf{Failure Analysis}. While our IRG-MotionLLM shows strong performance with extensive quantitative and qualitative results, it may fail on highly complex the goal text. \cref{fig:vis_appendix_failure} shows one of these cases. The goal text requires the model to generate a motion sequence including four sub-actions in a specific order. As shown in the upper part of \cref{fig:vis_appendix_failure}, although our IRG-MotionLLM obtains a three-round refined motion which is much more aligned with the goal than previous generated motions, it still misses the action of ``\emph{scratching the head}''. However, we note that generating perfect motion to such complex goal text is also challenging for existing advanced methods \cite{momask, mardm, motionagent}.


\section{Limitations and Future works}
\label{sec:limitations}
\noindent\textbf{IRMoGen with Advanced MoLLM Frameworks}. In this work, we take the first step to explore the novel IRMoGen paradigm within a general VQ-based motion-aware LLM framework. While this architecture elegantly supports IRMoGen via the LLM's autoregressive nature and enables fair comparisons with existing MoLLMs, it still has limitations. For example, the vanilla VQVAE may lose fine-grained motion details. To address this, {building} IRG-MotionLLM with advanced VQVAEs (\eg, RVQ \cite{momask,bu2025omnimogen}) can be valuable extension.
On the other hand, a concurrent work MotionGPT3 \cite{motiongpt3} adopts a Mix-of-Transformers \cite{mot} architecture to learn modality-aware knowledge with separate branches, so that the gap between symbolic sequences and continuous trajectories can be bridged. We believe exploring IRMoGen on such novel architectures is also a valuable direction for future research.

\vspace{0.2cm}
\noindent\textbf{Motion Assessment beyond Text-Motion Alignment}. Our primary focus is on linking generation with motion assessment and refinement to enhance alignment between motions and goal texts, and our experiments demonstrate the method's effectiveness. However, beyond text-motion alignment, aspects like physical plausibility and smoothness are also crucial for motion generation. Recently, MotionCritic \cite{motioncritic} proposes to train an additional model to provide quantitative scores for motion realism. However, how to endow MoLLMs with native physical realism assessment to optimize the IRMoGen process remains a worthwhile avenue for exploration. Also, introducing rewards related to physical realism during the Stage-3 training of IRG-MotionLLM is a meaningful future endeavor.

\vspace{0.2cm}
\noindent\textbf{Scaling Up IRG-MotionLLM}. We build our IRG-MotionLLM using a 2B LLM and conduct extensive experiments on the popular motion-text datasets, HumanML3D \cite{guo} and KIT-ML \cite{kit-ml}, to thoroughly validate the model's efficacy. Nevertheless, the limited datasets and model sizes may constrain generalization. We believe scaling up IRG-MotionLLM with larger foundation LLMs and training on larger text-motion datasets (\eg, MotionMillion \cite{go-to-zero}, ViMoGen \cite{lin2025quest}) will further enhance its interleaved reasoning abilities across a wider variety of goal texts. We leave these explorations for future work.

\vspace{0.2cm}
\noindent\textbf{Interleaved Reasoning for broader Motion-centric Tasks}. In this work, we explore whether interleaved reasoning could benefit motion generation. 
Taken T2M, the most general and representative task, as the testbed, we address the core challenges and analyze the properties of our model.
While text-motion interleaved reasoning for broader motion-centric tasks (\eg, motion editing, motion completion) still remains unexplored, our work provides an extensible technical route for future explorations. 
An intuitive extension of our work is to: (1) unify annotations of various motion-centric tasks and build dynamic motion-text interleaved reasoning traces; (2) adapt our multi-stage training scheme to train the model supporting various types of motion-text interleaved reasoning.

\clearpage
\newpage

\vspace{-0.2cm}

\clearpage
\newpage

\end{document}